\documentclass[journal]{IEEEtran}
\usepackage{latexsym,,amssymb,amsmath,graphicx,epsf,cite,bbm,float,subfig}
\usepackage{ifpdf}
\usepackage{epstopdf}

\AtBeginDocument{
\addtolength{\abovedisplayskip}{-0.3ex}
\addtolength{\abovedisplayshortskip}{-0.3ex}
\addtolength{\belowdisplayskip}{-0.3ex}
\addtolength{\belowdisplayshortskip}{-0.3ex}
}

\renewcommand{\vec}[1]{\mbox{$\mathbf{#1}$}}
\newcommand{\vw}{\vec{w}}
\newcommand{\vq}{\vec{q}}
\newcommand{\vtw}{\tilde{\vec{w}}}
\newcommand{\vx}{\vec{x}}

\newcommand{\mR}{\vec{R}}
\newcommand{\mQ}{\vec{Q}}
\newcommand{\mI}{\vec{I}}
\newcommand{\mSig}{\vec{\Sigma}}

\newcommand{\Rp}{\mbox{$\mathbbm{R}^p$}}
\newcommand{\R}{\mbox{$\mathbbm{R}$}}
\newcommand{\defi}{\stackrel{\bigtriangleup}{=}}
\newcommand{\nn}{\nonumber}
\newcommand{\Tr}{\mathrm{Tr}}
\newcommand{\sign}{\mathrm{sign}}

\begin{document}

\title{A Novel Family of Adaptive Filtering Algorithms Based on The Logarithmic Cost}
\author{Muhammed O. Sayin, N. Denizcan Vanli, Suleyman S. Kozat*,~{\em Senior Member, IEEE}
\thanks{This work is in part supported by the Outstanding Researcher Programme of Turkish Academy of Sciences and TUBITAK project 112E161.}
\thanks{The authors are with the Department of Electrical and Electronics Engineering, Bilkent University, Bilkent, Ankara 06800 Turkey, Tel: +90 (312) 290-2336, Fax: +90 (312) 290-1223 (e-mail: sayin@ee.bilkent.edu.tr, vanli@ee.bilkent.edu.tr, kozat@ee.bilkent.edu.tr).}}

\maketitle
\begin{abstract}
We introduce a novel family of adaptive filtering algorithms based on
a relative logarithmic cost. The new family intrinsically combines the
higher and lower order measures of the error into a single continuous
update based on the error amount. We introduce important members of
this family of algorithms such as the least mean logarithmic square
(LMLS) and least logarithmic absolute difference (LLAD) algorithms
that improve the convergence performance of the conventional
algorithms. However, our approach and analysis are generic such that
they cover other well-known cost functions as described in the
paper. The LMLS algorithm achieves comparable convergence performance
with the least mean fourth (LMF) algorithm and extends the stability
bound on the step size. The LLAD and least mean square (LMS)
algorithms demonstrate similar convergence performance in impulse-free
noise environments while the LLAD algorithm is robust against
impulsive interferences and outperforms the sign algorithm (SA). We
analyze the transient, steady state and tracking performance of the
introduced algorithms and demonstrate the match of the theoretical
analyzes and simulation results. We show the extended stability bound
of the LMLS algorithm and analyze the robustness of the LLAD algorithm
against impulsive interferences. Finally, we demonstrate the
performance of our algorithms in different scenarios through numerical
examples.
\end{abstract}

\begin{center}
\bfseries EDICS Category: MLR-LEAR, ASP-ANAL, MLR-APPL
\end{center}

\section{Introduction}
\IEEEPARstart{A}{daptive} filtering applications such as channel
equalization, noise removal or echo cancellation utilize a certain
statistical measure of the error signal\footnote{Time index appears as
  a subscript.} $e_t$ denoting the difference between the desired
signal $d_t$ and the estimation output $\hat{d}_t$. Usually, the mean
square error $E[e^2_t]$ is used as the cost function due to its
mathematical tractability and relative ease of analysis. The least
mean square (LMS) and normalized least mean square (NLMS) algorithms
are the members of this class~\cite{sayed_book}. In the literature,
different powers of the error are commonly used as the cost function
in order to provide stronger convergence or steady-state performance
than the least-squares algorithms under certain
settings~\cite{sayed_book}.

The least mean fourth (LMF) algorithm and its family use the even powers of the error as the cost function, i.e., $E[e^{2n}_t]$~\cite{walach1984}. This family achieves a better trade-off between the transient and steady-state performance, however, has stability issues~\cite{nascimento2005conf,nascimento2006,hubscher2007}. The stability of the LMF algorithm depends on the input and noise power, and the initial value of the adaptive filter weights~\cite{eweda2012}. On the other hand, the stability of the conventional LMS algorithm depends only on the input power for a given step-size~\cite{sayed_book}. The normalized filters improve the performance of the algorithms under certain settings by removing dependency to the input statistics in the updates~\cite{nascimento2004}. However, note that the normalized least mean fourth (NLMF) algorithm does not solve the stability issues~\cite{eweda2012}. In~\cite{eweda2012}, authors propose the stable NLMF algorithm, which might also be derived through the proposed relative logarithmic error cost framework as shown in this paper.

The performance of the least-squares algorithms degrades severely when
the input and desired signal pairs are perturbed by heavy tailed
impulsive interferences, e.g., in applications involving high power
noise signals~\cite{shao1993}. In this context, we define
\emph{robustness} as the insensitivity of the algorithms against the
impulsive interferences encountered in the practical applications and
provide a theoretical framework~\cite{kim1995}. Note that, usually,
the algorithms using lower-order measures of the error in their cost
function are relatively less sensitive to such perturbations. For
example, the well-known sign algorithm (SA) uses the $L_1$ norm of the
error and is robust against impulsive interferences since its update
involves only the sign of $e_t$. However, the SA usually exhibits
slower convergence performance especially for highly correlated input
signals~\cite{mathews1987}.

The mixed-norm algorithms minimize a combination of different error norms in order to achieve improved convergence performance~\cite{chambers1994,chambers1997}. For example,~\cite{chambers1997} combines the robust $L_1$ norm and the more sensitive but better converging $L_2$ norm through a mixing parameter. Even though the combination parameter brings in an extra degree of freedom, the design of the mixed norm filters requires the optimization of the mixing parameter based on a priori knowledge of the input and noise statistics. On the other hand, the mixture of experts algorithms adaptively combine different algorithms and provide improved performance irrespective of the environment statistics~\cite{garcia2003,garcia2005,garcia2006,silva2008}. However, note that such mixture approaches require to operate several different algorithms on parallel, which may be infeasible in different applications~\cite{kozat2010}. In~\cite{garcia2005imp}, authors propose an adaptive combination of $L_1$ and $L_2$ norms of the error in parallel, however, the resulting algorithm demonstrates impulsive perturbations on the learning curves. This results since the impulsive interferences severely degrade the algorithmic updates. In general, the samples contaminated with impulses contain little useful information~\cite{kim1995}. Hence, the robust algorithms need to be less sensitive only against large perturbations on the error and can be as sensitive as the conventional least squares algorithms for small error values. The switched-norm algorithms switch between the $L_1$ and $L_2$ norms based on the error amount such as the robust Huber filter~\cite{petrus1999}. This approach combines the better convergence of $L_2$ and the robustness of $L_1$ together in a discrete manner with a breaking point in the cost function, however, requires optimization of certain parameters as detailed in this paper.

In this paper, we use \emph{diminishing return} functions, e.g., the
logarithm function, as a normalization (or a regularization) term,
i.e., as a subtracting term, in the cost function in order to improve
the convergence performances. We particularly choose the logarithm
function as the normalizing diminishing return function \cite{realbook}
in our cost definitions since the logarithmic function is
differentiable and results efficient and mathematically tractable
adaptive algorithms. As shown in the paper, by using the logarithm
function, we are able to use of the higher-order statistics of the
error for small perturbations. On the other hand, for larger error
values, the introduced algorithms seek to minimize the conventional
cost functions due to the decreasing weight of the logarithmic term
with the increasing error amount. In this sense, the new framework is
akin to a continuous generalization of the switched norm algorithms,
hence greatly improve the convergence performance of the mixed-norm
methods as shown in this paper.

Our main contributions include: 1) We propose the least mean
logarithmic square (LMLS) algorithm, which achieves a similar
trade-off between the transient and steady-state performance of the
LMF algorithm and as stable as the LMS algorithm; 2) We propose the
least logarithmic absolute difference (LLAD) algorithm, which
significantly improves the convergence performance of the SA while
exhibiting comparable performance with the SA in the impulsive noise
environments; 3) We analyze the transient, steady-state and tracking
performance of the introduced algorithms; 4) We demonstrate the
extended stability bound on the step-sizes with the logarithmic error
cost framework; 5) We introduce an impulsive noise framework and
analyze the robustness of the LLAD algorithm in the impulsive noise
environments; 6) We demonstrate the significantly improved convergence
performances of the introduced algorithms in several different
scenarios in our simulations.

We organize the paper as follows. In Section II, we introduce the
relative logarithmic error cost framework. In Section III, the
important members of the novel family are derived. We analyze the
transient, steady-state and tracking performances of those members in
Section IV. In Section V, we compare the stability bound on the
step-sizes and the robustness of the proposed algorithms. In Section
VI, we provide the numerical examples demonstrating the improved
performance of the conventional algorithms in the new logarithmic
error cost framework. We conclude the paper in Section VII with
several remarks.\\

\noindent
{\bf Notation:} Bold lower (or upper) case letters denote the vectors
(or matrices). For a vector $\vec{a}$ (or matrix $\vec{A}$),
$\vec{a}^T$ (or $\vec{A}^T$) is its ordinary transpose. $\|\cdot\|$ and
$\|\cdot\|_{\vec{A}}$ denote the $L_2$ norm and the weighted $L_2$ norm with the matrix $\vec{A}$, respectively (provided that $\vec{A}$
is positive-definite). $|\cdot|$ is the absolute value operator. We
work with real data for notational simplicity. For a random variable
$x$ (or vector $\vx$), $E[x]$ (or $E[\vx]$) represents its
expectation. Here, $\Tr(\vec{A})$ denotes the trace of the matrix
$\vec{A}$ and $\nabla_{\vec{x}}f(\vec{x})$ is the gradient operator.

\begin{figure}[t!]
\centering
\includegraphics[width=2.5in]{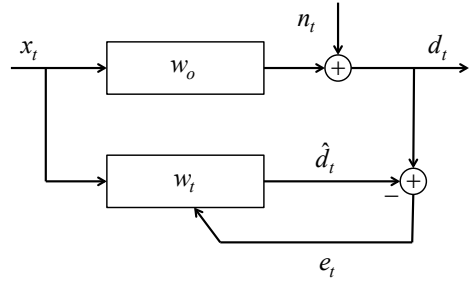}
\caption{General system identification configuration.}
\label{fig:model}
\end{figure}

\section{Cost Function With Logarithmic Error}
We consider the system identification framework shown in Fig.~\ref{fig:model}, where we denote the input signal by $\vx_t$ and the desired signal by $d_t$. Here, we observe an unknown vector\footnote{Although we assume a time invariant unknown system vector here, we also provide the tracking performance analysis for certain non-stationary models later in the paper.} $\vw_o \in \Rp$ through a linear model
\begin{align*}
d_t= \vw_o^T\vx_t + n_t,
\end{align*}
where $n_t$ represents the noise and we define the error signal as $e_t \defi d_t - \hat{d}_t = d_t - \vw^T_t\vx_t$. In this framework, adaptive filtering algorithms estimate the unknown system vector $\vw_o$ through the minimization of a certain cost function. The gradient descent methods usually employ convex and uni-modal cost functions in order to converge to the global minimum of the error surfaces, e.g., the mean square error $E[e^2_t]$~\cite{sayed_book}. The different powers of $e_t$~\cite{walach1984, mathews1987} or a linear combination of different error powers~\cite{chambers1994,chambers1997} are also widely used.

\begin{figure}[t!]
\centering
\includegraphics[width=3in]{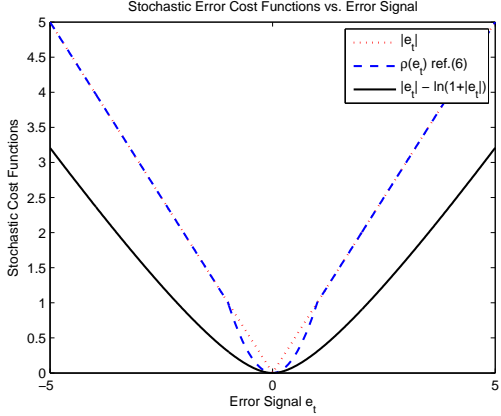}
\caption{Here, we plot stochastic cost functions to illustrate decreased steepness of the least squares algorithms in the logarithmic error cost framework for small error amounts.}
\label{fig:cost}
\end{figure}

In this framework, we use a normalized error cost function using the logarithm function given by
\begin{align}
J\left(e_t\right) \defi F\left(e_t\right) - \frac{1}{\alpha} \ln \left(1+\alpha F\left(e_t\right)\right), \label{eq:cost}
\end{align}
where $\alpha > 0$ is a design parameter and $F\left(e_t\right)$ is a conventional cost function of the error signal $e_t$, e.g., $F(e_t) = E\left[|e_t|\right]$. As an illustration, in Fig.~\ref{fig:cost}, we compare $|e_t|$ and $|e_t|-\ln(1+|e_t|)$. From this plot, we observe that the  logarithm based cost function is less steep for small perturbations on the error while both logarithmic square and absolute difference cost functions exhibit comparable steepness for large error values. Indeed, this new family intrinsically combines the benefits of using lower and higher-order measures of the error into a single adaptation algorithm. Our algorithms provide comparable convergence rate with a conventional algorithm minimizing the cost function $F(e_t)$ and achieve smaller steady-state mean square errors through the use of  higher-order statistics for small perturbations of the error.\\

\noindent
{\bf Remark 2.1:}
In~\cite{song2013}, the authors propose a stochastic cost function using the logarithm function as follows
\begin{align*}
J_{\mbox{\cite{song2013}}}(e_t) \defi \frac{1}{2\gamma}\ln\left(1+\gamma\left(\frac{e_t}{\|\vx_t\|}\right)^2\right).
\end{align*}
Note that the cost function $J_{\mbox{\cite{song2013}}}(e_t)$ is the
subtracted term in~\eqref{eq:cost} for $F(e_t) =
\frac{e^2_t}{\|\vx_t\|^2}$. The Hessian matrix of
$J_{\mbox{\cite{song2013}}}(e_t)$ is given by
\begin{align*}
\vec{H}\left(J_{\mbox{\cite{song2013}}}(e_t)\right) = &\frac{\vx_t\vx_t^T}{\|\vx_t\|^2\left(1+\gamma\left(\frac{e_t}{\|\vx_t\|}\right)^2\right)}\\
&\times \left(1-\frac{2\gamma e_t^2}{\|\vx_t\|^2\left(1+\gamma\left(\frac{e_t}{\|\vx_t\|}\right)^2\right)}\right).
\end{align*}
We emphasize that $\vec{H}\left(J_{\mbox{\cite{song2013}}}(e_t)\right)$ is positive semi-definite provided that $\gamma\left(\frac{e_t}{\|\vx_t\|}\right)^2 \leq 1$, thus, the
parameter $\gamma$ should be chosen carefully to be able to
efficiently use the gradient descent algorithms. On the other hand, we
show that the new cost function in~\eqref{eq:cost} is a convex
function enabling the use of the diminishing return property~\cite{realbook} of
the logarithm function for stable and robust updates.\\

The relative logarithmic error cost we introduce in \eqref{eq:cost} can also be expressed as
\begin{align}
J(e_t) = \frac{1}{\alpha}\ln\left(\frac{\exp\left(\alpha F(e_t)\right)}{1+\alpha F(e_t)}\right).\label{eq:excost}
\end{align}
Since $\exp(\alpha F(e_t)) = \sum_{m=0}^{\infty} \frac{1}{m!}\alpha^m F^{m}(e_t)$, we obtain
\begin{align}
J(e_t) = \frac{1}{\alpha}\ln\left(1 + \frac{\frac{\alpha^2}{2!}F^2(e_t)+\frac{\alpha^3}{3!}F^3(e_t) + \cdots}{1+\alpha F(e_t)}\right).\label{eq:excost2}
\end{align}
Since $F(e_t)$ is a non-negative function, $J(e_t)$ is also a non-negative function by~\eqref{eq:excost2}.\\

\noindent
{\bf Remark 2.2:}
The Hessian matrix of $J(e_t)$ is given by
\begin{align*}
\vec{H}\left(J(e_t)\right) \hspace{-0.1cm} = \hspace{-0.1cm} \vec{H}\left(F(e_t)\right)\frac{\alpha F(e_t)}{1+\alpha F(e_t)} \hspace{-0.1cm} + \hspace{-0.1cm} \frac{\alpha\nabla_{\vw}F(e_t)\nabla_{\vw}F(e_t)^T}{\left(1+\alpha F(e_t)\right)^2},
\end{align*}
which is positive semi-definite provided that $\vec{H}\left(F(e_t)\right)$ is a positive semi-definite matrix.\\

We obtain the first gradient of~\eqref{eq:cost} as follows
\begin{align*}
\nabla_{\vw}J(e_t) = \nabla_{\vw}F(e_t)\frac{\alpha F(e_t)}{1+ \alpha F(e_t)},
\end{align*}
which yields zero if $\nabla_{\vw}F(e_t)$ or $F(e_t)$ is zero. Note that the optimal solution for the cost function $F(e_t)$ minimizes $F(e_t)$ and is obtained by
\begin{align*}
\nabla_{\vw = \vw_o}F(e_t) = 0.
\end{align*}
Since $F(e_t)$ is a non-negative convex function, the global minimum and the value yielding zero gradient coincide if the latter exits. Hence, the optimal solution for the relative logarithmic error cost function is the same with the cost function $F(e_t)$ since as shown in Remark 2.2 the Hessian matrix of the logarithmic cost function is positive semi-definite. For example, the mean-square error cost function $F(e_t) = E[e^2_t]$ yields to the Wiener solution $\vw_o = E[\vx_t\vx_t^T]^{-1}E[\vx_td_t]$.\\

\noindent
{\bf Remark 2.3:}
By Maclaurin series of the natural logarithm for $\alpha F(e_t) \leq 1$,~\eqref{eq:cost} yields
\begin{align}
J(e_t) &= F(e_t) - \frac{1}{\alpha}\left(\alpha F(e_t) - \frac{\alpha^2}{2}F^2(e_t) + \cdots\right)\nn\\
&= \frac{\alpha}{2}F^2(e_t) - \frac{\alpha^2}{3}F^3(e_t) + \cdots, \label{eq:cost2}
\end{align}
which is an infinite combination of the conventional cost function for small values of $F(e_t)$. We emphasize that the cost function~\eqref{eq:cost2} yields to the second power of the cost function $F(e_t)$ for small values of the error while for large error values, the cost function $J(e_t)$ resembles $F(e_t)$ as follows:
\begin{align*}
F(e_t) - \frac{1}{\alpha}\ln\left(1+\alpha F(e_t)\right) \rightarrow F(e_t)\;\mbox{as}\;e_t\rightarrow\infty.
\end{align*}
Hence, the new methods are the combinations of the algorithms with
mainly $F^2(e_t)$ or $F(e_t)$ cost functions based on the error
amount. It is important to note that the objective functions
$F^2(e_t)$, e.g., $E[e_t^2]^2,$ and $F(e_t^2)$, e.g., $E[e_t^4],$
yields the same stochastic gradient update after removing the
expectation  in this paper. The switched norm algorithms also
combine two different norms into a single update in a discrete manner
based on the error amount. As an example, the Huber objective function
combining $L_1$ and $L_2$ norms of the error is defined
as~\cite{petrus1999}
\begin{align}
\rho(e_t) \defi \left\{\begin{array}{ll} \frac{1}{2}e_t^2 & \mbox{for}\;|e_t| \leq \gamma, \\ \gamma|e_t| - \frac{1}{2}\gamma^2 & \mbox{for}\;|e_t| > \gamma, \end{array}\right.\label{eq:huber}
\end{align}
where $\gamma > 0$ denotes the cut-off value. In Fig.~\ref{fig:cost}, we also compare the Huber objective function (for $\gamma = 1$) and the introduced cost~\eqref{eq:cost} with $F(e_t)=E[|e_t|]$ (for $\alpha = 1$). Note that~\eqref{eq:huber} uses a piecewise-function combining two different algorithms based on the comparison of the error with the cut-off value $\gamma$. On the other hand, logarithm based cost function $J(e_t)$ intrinsically combines the functions with different order of powers in a continuous manner into a single update and avoids possible anomalies that might arise due to the breaking point in the cost function.\\

\noindent
{\bf Remark 2.4:} Instead of a logarithmic normalization term, it is
also possible to use various functions having diminishing returns property in order to
provide stability and robustness to the conventional algorithms. For
example, one can choose the cost function as
\begin{align}
J_{\mathrm{arctan}}(e_t) \defi F(e_t) - \frac{1}{\alpha}\arctan\left(\alpha F(e_t)\right)\label{eq:altcost}
\end{align}
and the Taylor series expansion of the second term in~\eqref{eq:altcost} around $F(e_t) = 0$ is given by
\begin{align*}
\frac{1}{\alpha}\arctan\left(1+\alpha F(e_t)\right) = F(e_t) - \frac{\alpha^2}{3}F^3(e_t) + \cdots.
\end{align*}
Thus, the resulting algorithm combines the algorithms using mainly $F^3(e_t)$ (for small perturbations on the error) and $F(e_t)$. We note that the algorithms using~\eqref{eq:altcost} are also as stable as $F(e_t)$, however, they behave like minimizing the higher-order measures, i.e., $F^3(e_t)$, for small error values.\\

In the next section, we propose important members of this novel adaptive filter family.\\

\section{Novel Algorithms}
Based on the gradient of $J(e_t)$ we obtain the general steepest descent update as
\begin{align*}
\vw_{t+1} = \vw_t - \mu\,\nabla_{\vw}F(e_t)\frac{\alpha F(e_t)}{1+\alpha F(e_t)},
\end{align*}
where $\mu > 0$ is the step size and $\alpha > 0$ is the design parameter. \\

\noindent
{\bf Remark 3.1:}
In the previous section, we motivate the logarithm based error cost framework as a continuous generalization of the switched norm algorithms. The switched norm update involves a cut-off $\gamma$ in the comparison of the error amount. Similarly, we utilize a design parameter $\alpha$ in~\eqref{eq:cost} in order to determine the asymptotic cut-off value. For example, a larger $\alpha$ decreases the weight of the logarithmic term in the cost~\eqref{eq:cost} and the resulting algorithm behaves more like minimizing the cost $F(e_t)$. In the performance analyzes, we show that a sufficiently small design parameter, i.e., $\alpha = 1$, does not have determinative influence on the steady-state convergence performance under the Gaussian noise signal assumption. Hence, in the following algorithms we choose $\alpha = 1$. On the other hand, we resort to the usage of $\alpha$ in order to facilitate the performance analyzes of the algorithms. Additionally, in the impulsive noise environments, we show that the optimization of $\alpha$ improves the steady-state convergence performance of the introduced algorithms.\\

If we assume that  after removing the expectation to generate stochastic gradient updates $F(e_t)$ yields $f(e_t)$, e.g., $F(e_t) = E[f(e_t)]$, then the general stochastic gradient update is given by
\begin{align}
\vw_{t+1}&=\vw_{t} - \mu\, \nabla_{\vw}e_t\nabla_{e_t}f(e_t)\frac{f(e_t)}{1+f(e_t)},\nn\\
&=\vw_t + \mu\, \vx_t\nabla_{e_t}f(e_t)\frac{f(e_t)}{1+f(e_t)}.\label{eq:update}
\end{align}

In the following subsections, we introduce algorithms improving the performance of the conventional algorithms such as the LMS (i.e. $f(e_t)=e^2_t$), sign algorithm (i.e. $f(e_t)=|e_t|$) and normalized updates.

\subsection{The Least Mean Logarithmic Square (LMLS) Algorithm}
For $F(e_t) = E[e^2_t]$, the stochastic gradient update yields
\begin{align}
\vw_{t+1} &= \vw_t + \mu \vx_t e_t \frac{e^2_t}{1+e^2_t}\nn\\
&= \vw_t + \mu \frac{\vx_te^3_t}{1+e^2_t}.\label{eq:LMLS}
\end{align}
Note that we include the multiplier `$2$' coming from the gradient $\nabla_{e_t}e^2_t = 2e_t$ into the step-size $\mu$. The algorithm~\eqref{eq:LMLS} resembles a least-mean fourth update for the small error values while it behaves like the least-mean square algorithm for large perturbations on the error. This provides smaller steady-state mean square error thanks to the fourth-order statistics of the error for small perturbations and stability of the least-squares algorithms for large perturbations. Hence, the LMLS algorithm intrinsically combines the least mean-square and least-mean fourth algorithms based on the error amount instead of mixed LMF + LMS algorithms \cite{chambers1994} that need artificial combination parameter in the cost definition.

\subsection{The Least Logarithmic Absolute Difference (LLAD) Algorithm}
The SA utilizes $F(e_t) = E[|e_t|]$ as the cost function, which provides robustness against impulsive interferences~\cite{sayed_book}. However, the SA has slower convergence rate since the $L_1$ norm is the smallest possible error power for a convex cost function. In the logarithmic cost framework, for $F(e_t) = E[|e_t|]$,~\eqref{eq:update} yields
\begin{align}
\vw_{t+1} &= \vw_t + \mu \vx_t\sign (e_t)\frac{|e_t|}{1+|e_t|}\nn\\
&= \vw_t + \mu\frac{\vx_te_t}{1+|e_t|}.\label{eq:LLAD}
\end{align}
The algorithm~\eqref{eq:LLAD} combines the LMS algorithm and SA into a single robust algorithm with improved convergence performance. We note that in Section V we calculate the optimum $\alpha_{\mathrm{opt}}$ in order to achieve better convergence performance than the SA in the impulsive noise environments.

\subsection{Normalized Updates}
We introduce normalized updates with respect to the regressor signal in order to provide independence from the input data correlation statistics under certain settings. We define the new objective function as
\begin{align*}
J_{\mathrm{new}} (e_t) \defi F\left(\frac{e_t}{\|\vx_t\|}\right) - \frac{1}{\alpha}\ln\left(1+\alpha F\left(\frac{e_t}{\|\vx_t\|}\right)\right),
\end{align*}
for example $F\left(\frac{e_t}{\|\vx_t\|}\right) = E\left[\frac{e_t^2}{\|\vx_t\|^2}\right]$. The Hessian matrix of the new cost function $J_{\mathrm{new}}(e_t)$ is also positive semi-definite provided that the Hessian matrix of $F\left(\frac{e_t}{\|\vx_t\|}\right)$ is positive semi-definite as shown in Remark 2.2.

The steepest-descent update is given by
\begin{align*}
\vw_{t+1} = \vw_t - \mu \nabla_{\vw}F\left(\frac{e_t}{\|\vx_t\|}\right)\frac{\alpha F\left(\frac{e_t}{\|\vx_t\|}\right)}{1 + \alpha F\left(\frac{e_t}{\|\vx_t\|}\right)}.
\end{align*}
For $F(\frac{e_t}{\|\vx_t\|}) = E\left[\right(\frac{e_t}{\|\vx_t\|}\left)^2\right]$, we get the normalized least mean logarithmic square (NLMLS) algorithm given by
\begin{align}
\vw_{t+1} = \vw_t + \mu \frac{\vx_{t}e^3_{t}}{\|\vx_{t}\|^2\left(\|\vx_{t}\|^2+e^2_{t}\right)}.\label{eq:NLMLS}
\end{align}
We point out that~\eqref{eq:NLMLS} is also proposed as the stable normalized least mean-fourth algorithm in~\cite{eweda2012}.

For $F(\frac{e_t}{\|\vx_t\|}) = E\left[\frac{|e_t|}{\|\vx_t\|}\right]$, we obtain the normalized least logarithmic absolute difference (NLLAD) algorithm as
\begin{align*}
\vw_{t+1} = \vw_t + \frac{\mu\, \vx_te_t}{\|\vx_t\|\left(\|\vx_t\|+|e_t|\right)}.
\end{align*}

In the next section, we analyze the transient and steady state performance of the introduced algorithms.

\section{Performance Analysis}
We define \emph{a priori} estimation error and the weighted form as
\begin{align*}
e_{a,t} \defi \vx^T_t\vtw_t\;\mbox{and}\;e_{a,t}^{\mSig}\defi \vx^T_t\mSig\vtw_t,
\end{align*}
where $\vtw_t \defi \vw_o - \vw_t$ and $\mSig$ is a symmetric positive definite weighting matrix. Different choice of $\mSig$ leads to the different performance measures of the algorithm~\cite{sayed_book}. In the analyzes, we include the design parameter $\alpha$ in order to facilitate the theoretical analyzes. After some algebra, we obtain the weighted-energy recursion~\cite{sayed_book,norma2003,tareq2003} as
\begin{align}
E\left[\|\vtw_{t+1}\|_{\mSig}^2\right]& = E\left[\|\vtw_t\|_{\mSig}^2\right]\nn \\
-\mu 2& E\left[e_{a,t}^{\mSig}\nabla_{e_t}f(e_t)\frac{\alpha f(e_t)}{1+\alpha f(e_t)}\right]\nn\\
+\mu^2& E\left[\|\vx_t\|_{\mSig}^2\left(\nabla_{e_t}f(e_t)\frac{\alpha f(e_t)}{1+\alpha f(e_t)}\right)^2\right].\label{energy1}
\end{align}
For notational simplicity, we define
\begin{equation}\label{eq:g_func}
  g(e_t) \defi \nabla_{e_t}f(e_t)\frac{\alpha f(e_t)}{1+\alpha f(e_t)}.
\end{equation}
Then,~\eqref{energy1} yields the general weighted-energy recursion~\cite{tareq2003} as follows
\begin{align}\label{eq:weightedenergyrec}
E\left[\|\vtw_{t+1}\|_{\mSig}^2\right] = &\,E\left[\|\vtw_t\|_{\mSig}^2\right] - \mu 2 E\left[e_{a,t}^{\mSig}g(e_t)\right]\nn\\
&+\mu^2E\left[\|\vx_t\|_{\mSig}^2g^2(e_t)\right].
\end{align}

In the subsequent analysis of~\eqref{energy1}, we use the following assumptions:

\begin{description}
  \item[{\it Assumption 1:}] \hfill \\
The observation noise $n_t$ is zero-mean independently and identically distributed (i.i.d.) Gaussian random variable and independent from $\vx_t$. The regressor signal $\vx_t$ is also zero-mean i.i.d. Gaussian random variable with the auto-correlation matrix $\mR \defi E\left[\vx_t\vx^T_t\right]$.
  \item[{\it Assumption 2:}] \hfill \\
The estimation error $e_t$ and the noise $n_t$ are jointly Gaussian. The Gaussian estimation error assumption is acceptable for sufficiently small step size $\mu$ and through the Assumption 1~\cite{sayed_book}.
  \item[{\it Assumption 3:}] \hfill \\
The estimation error $e_{t}$ is jointly Gaussian with the weighted \emph{a priori} estimation error $e_{a,t}^{\mSig}$ for any constant matrix $\mSig$. The assumption is reasonable for long filters, i.e. $p$ is large, sufficiently small step size $\mu$~\cite{tareq2003}, and by Assumption 2.
  \item[{\it Assumption 4:}] \hfill \\
The random variables $\|\vx_t\|_{\mSig}^2$ and $g^2(e_t)$ are uncorrelated, which enables the following split as
\begin{align*}
E\left[\|\vx_t\|_{\mSig}^2g^2(e_t)\right] = E\left[\|\vx_t\|_{\mSig}^2\right]E\left[g^2(e_t)\right].
\end{align*}
\end{description}

We next analyze the transient behavior of the new algorithms through the energy recursion \eqref{energy1}.

\subsection{Transient Analysis}
In the following we evaluate~\eqref{energy1} term by term. We first consider the second term in the right hand side (RHS) of \eqref{eq:weightedenergyrec} and introduce the following lemma

{\bf Lemma 1:}
{\em Under Assumptions 1-4, we have
\begin{equation}\label{eq:price}
  E[e_{a,t}^{\mSig}g(e_t)] = E[e_{a,t}^{\mSig}e_t]\frac{E[e_tg(e_t)]}{E[e^2_t]}.
\end{equation}
}

{\em Proof:}
The proof of Lemma 1 follows from the Price's result \cite{price1958,mcmahon1964}. That is, for any Borel function $g(b)$ we can write
\begin{align*}
E[xg(y)] = \frac{E[xy]}{E[y^2]}E[yg(y)],
\end{align*}
where $x$ and $y$ are zero-mean jointly Gaussian random variables~\cite{koh1985}. Hence by Assumptions 2 and 3, we obtain \eqref{eq:price} and the proof is concluded. \hfill $\square$

Since $e_t = e_{a,t} + n_t$, we obtain
\begin{align}
E\left[e_{a,t}^{\mSig}e_t\right] = E\left[e_{a,t}^{\mSig}e_{a,t}\right] = E\left[\|\vtw_t\|_{\mSig\vx_t\vx^T_t}^2\right],\label{eq:2}
\end{align}
by Assumption 1. Additionally, by the independence assumption for the regressor $\vx_t$ (i.e., Assumptions 1 and 4), we can simplify the third term in the RHS of \eqref{eq:weightedenergyrec}. Hence, the weighted-error recursion \eqref{eq:weightedenergyrec} could be written as follows \cite{tareq2003}
\begin{align}
E\left[\|\vtw_{t+1}\|_{\mSig}^2\right] = \,&E\left[\|\vtw_t\|_{\mSig}^2\right] -\mu2h_G\left(e_t\right)E\left[\|\vtw_t\|_{\mSig\mR}^2\right]\nn\\
&+ \mu^2E\left[\|\vx_t\|_{\mSig}^2\right]h_U\left(e_t\right),\label{eq:recursion}
\end{align}
where
\begin{align*}
h_G(e_t)\defi \frac{E[e_tg(e_t)]}{E[e^2_t]}, \;\; h_U(e_t)\defi E\left[g^2(e_t)\right].
\end{align*}

\begin{table*}[!t]
\renewcommand{\arraystretch}{2}
\caption{$h_G(e_t)$ and $h_U(e_t)$ corresponding to the stochastic costs $e^2_t$ and $|e_t|$, where $\sigma_e^2 = E[e^2_t]$ and $\lambda = \frac{1}{2\alpha\sigma_e^2} = \alpha\kappa$.}
\label{hgu}
\begin{center}
  \begin{tabular}{ c | c | c}
    \hline
    Algorithm & $h_G(e_t)$ & $h_U(e_t)$\\
    \hline
    LMF & $3 \sigma_e^2$ & $15 \sigma_e^6$\\
    \hline
    LMLS & $1 - 2\lambda\left(1-\sqrt{\pi\lambda}\mathrm{exp}(\lambda)\mathrm{erfc}(\sqrt{\lambda})\right)$ &  $\sigma_e^2\left(1-2\lambda ( \lambda + 2) + \lambda (2 \lambda + 5) \sqrt{\pi\lambda}\mathrm{exp}(\lambda)\mathrm{erfc}\left( \sqrt{\lambda} \right)\right)$ \\
    \hline
    LMS & 1 & $\sigma_e^2$\\ 	
    \hline
    LLAD & $\frac{1}{\sigma_e}\sqrt{\frac{2}{\pi}}\left(1-\sqrt{\kappa\pi} + \kappa\frac{\pi\mathrm{erfi}(\sqrt{\kappa})-\mathrm{Ei}(\kappa)}{\mathrm{exp}(\kappa)}\right)$ & $1 - 2\kappa + 2 \sqrt{\frac{\kappa}{\pi}} \left( 1 + (\kappa-1) \frac{\pi \mathrm{erfi}\left(\sqrt{\kappa}\right)-\mathrm{Ei}(\kappa)}{\mathrm{exp}(\kappa)} \right)$\\
    \hline
    SA & $\frac{1}{\sigma_e}\sqrt{\frac{2}{\pi}}$ & $1$\\
    \hline
  \end{tabular}
\end{center}
\end{table*}

\noindent {\bf Remark 4.1: }In the Appendices we evaluate the functions $h_G(e_t)$ and $h_U(e_t)$ for the LMLS and LLAD algorithms and tabulate the evaluated results with the results for the LMS algorithm, LMF algorithm and SA in Table~\ref{hgu}.

Using~\eqref{eq:recursion}, in the following we construct the learning curves for the new algorithms:

{\em i)} For the white regression data for which $\mR = \sigma_x^2\mI$, the time-evolution of the mean square deviation (MSD) $E[\|\vtw_t\|^2]$ is given by
\begin{align*}
E\left[\|\vtw_{t+1}\|^2\right] \hspace{-0.1cm} = \hspace{-0.1cm} \left(1- \mu2\sigma_x^2h_G(e_t)\right)E\left[\|\vtw_t\|^2\right] \hspace{-0.1cm} + \hspace{-0.1cm} \mu^2p\sigma_x^2h_U(e_t).
\end{align*}
This completes the transient analysis of the MSD for the white regressor data since $h_U(e_t)$ and $h_G(e_t)$ are given in Table~\ref{hgu}, and the right hand side only depends on $E[\|\vtw_t\|^2]$.

{\em ii)} For the correlated regression data, by the Cayley-Hamilton theorem after some algebra we get the state-space recursion
\begin{align*}
{\cal W}_{t+1} = {\cal A} {\cal W}_t + \mu^2 \cal Y
\end{align*}
where the vectors are defined as
\begin{align*}
{\cal W}_t \defi \begin{bmatrix} E\left[\|\vtw_t\|^2\right] \\ \vdots\\ E\left[\|\vtw_t\|_{\mR^{p-1}}^2\right]\end{bmatrix}, \;{\cal Y} \defi h_U(e_t)\begin{bmatrix}E\left[\|\vx_t\|^2\right]\\ \vdots \\ E\left[\|\vx_t\|_{\mSig^{p-1}}^2\right]\end{bmatrix}.
\end{align*}
The coefficient matrix $\cal A$ is given by
\begin{align*}
{\cal A} \defi \begin{bmatrix} 1 & -2\mu h_G(e_t) & \cdots & 0 \\ 0 & 1 & \cdots & 0 \\ \vdots & \vdots & \ddots & \vdots \\ 2\mu c_0 h_G(e_t) & 2 \mu c_1 h_G(e_t) & \cdots & 1 + 2\mu c_{p-1}h_G(e_t) \end{bmatrix}.
\end{align*}
where the $c_i$'s for $i \in \{0,1,...,p-1\}$ are the coefficients of the characteristic polynomial of $\mR$. Note that the top entry of the state vector ${\cal W}_t$ yields the time-evolution of the mean square deviation $E\left[\|\vtw_t\|^2\right]$ and the second entry gives the learning curves for the excess mean square error $E\left[e_{a,t}^2\right]$.

In the following subsection, we analyze the steady state excess mean square error (EMSE) and MSD of the LMLS and LLAD algorithms.

\begin{figure*}[!t]
\centering
\subfloat[The LMLS Algorithm]{\includegraphics[width=3.3in]{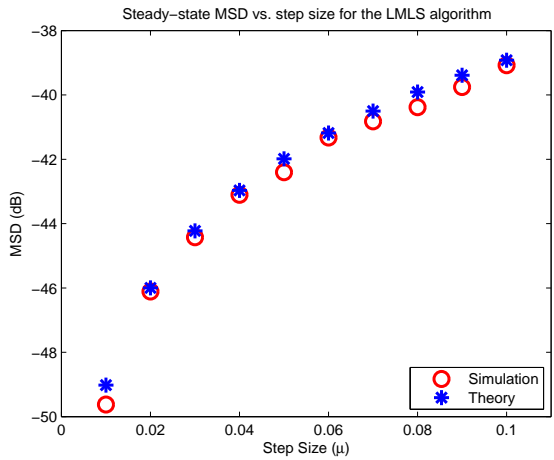}}
\hfil
\subfloat[The LLAD Algorithm]{\includegraphics[width=3.3in]{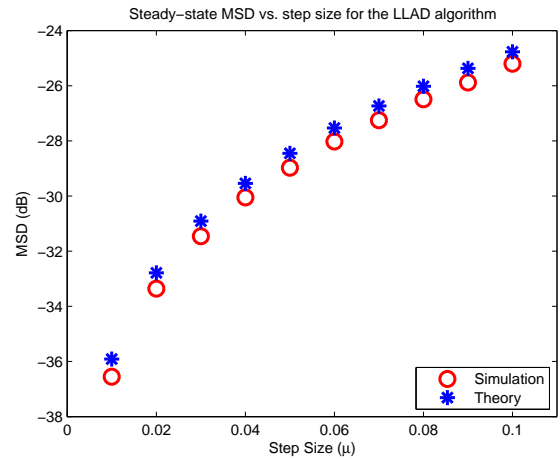}}
\caption{Dependence of the steady-state MSD on the step size $\mu$ for the LMLS and LLAD algorithms.}
\label{fig_steady}
\end{figure*}

\begin{figure*}[!t]
\centering
\subfloat[MSD]{\includegraphics[width=3.3in]{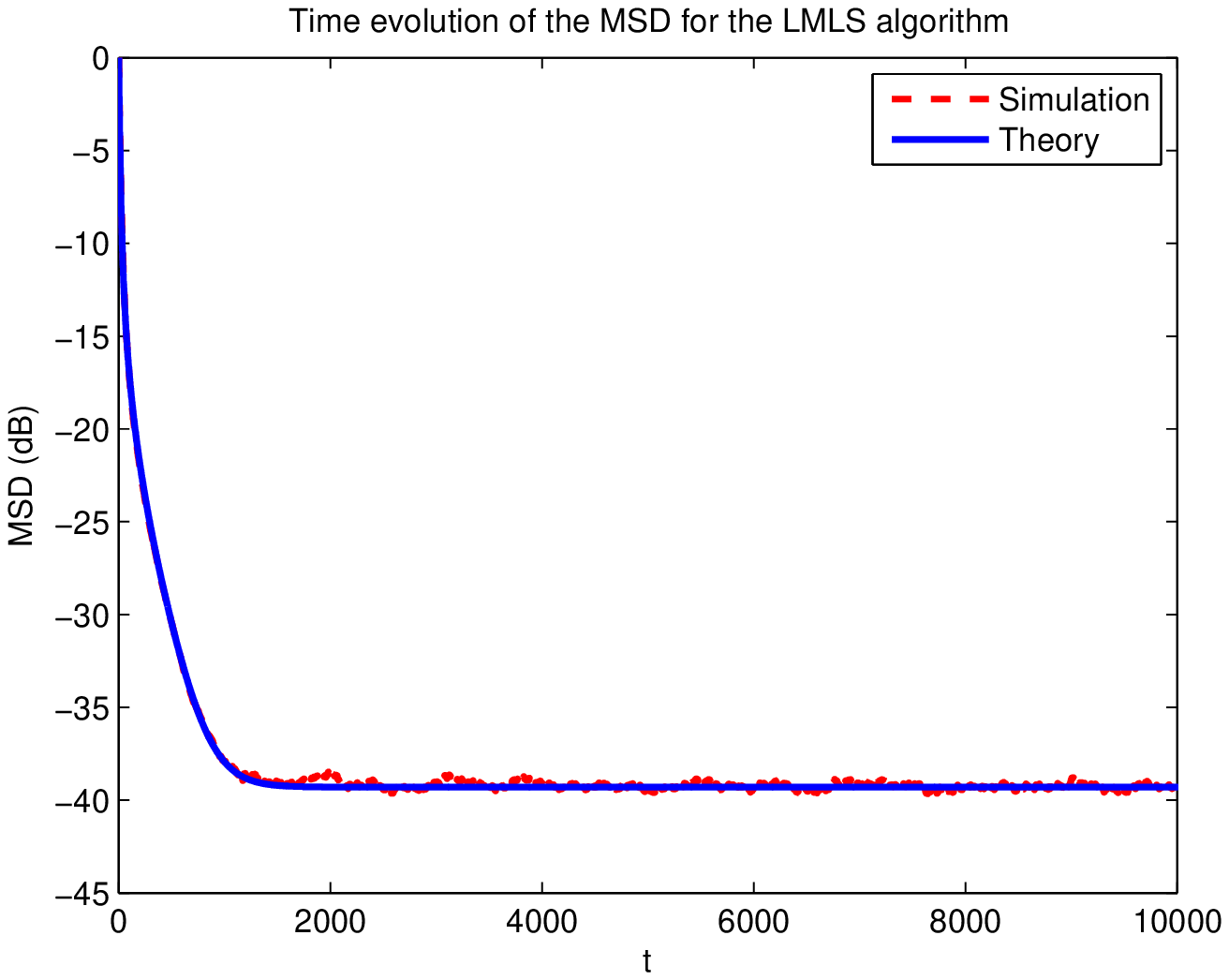}}
\hfil
\subfloat[EMSE]{\includegraphics[width=3.3in]{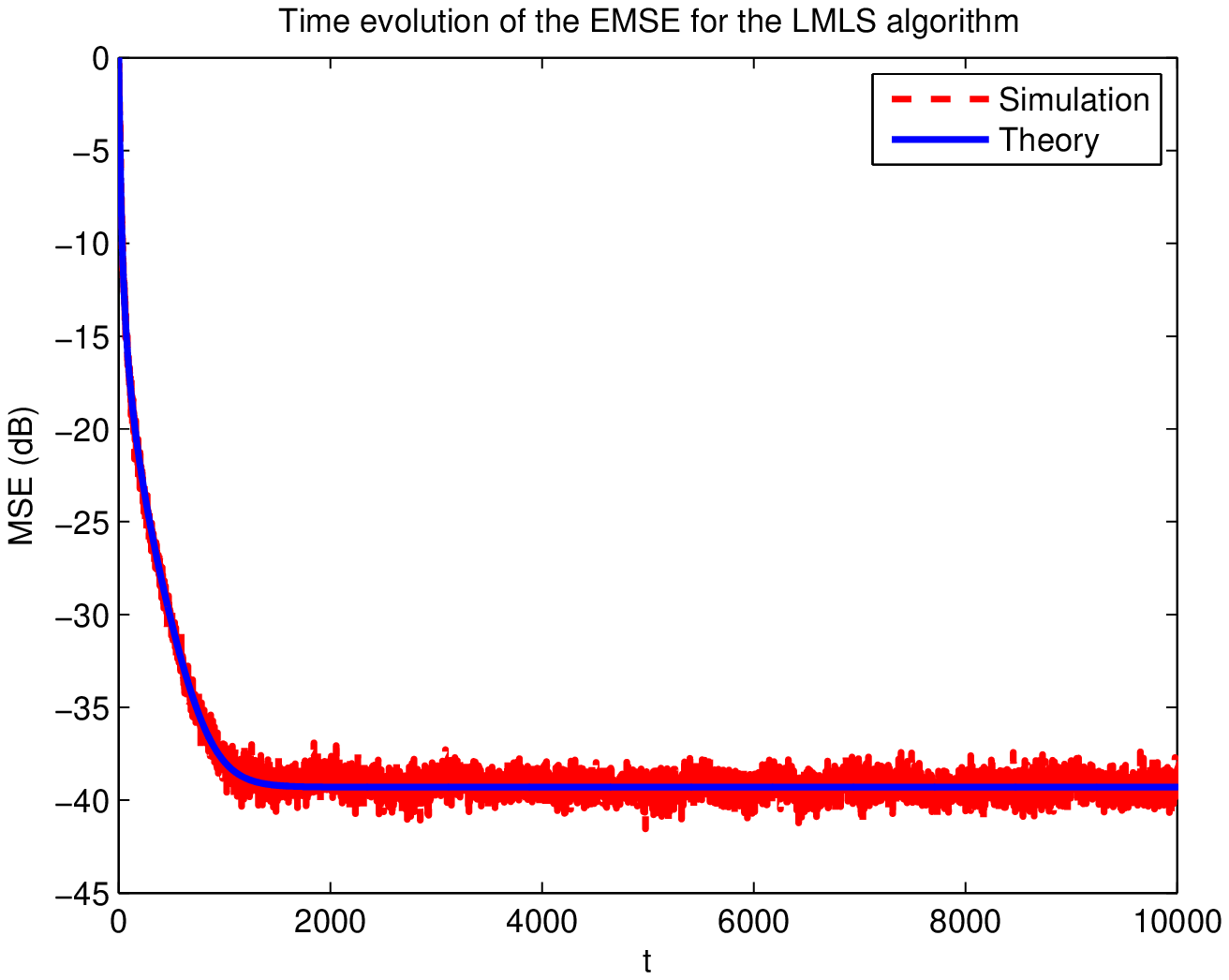}}
\caption{Theoretical and simulated MSD and EMSE for the LMLS algorithm.}
\label{fig_transient_LMLS}
\end{figure*}

\begin{figure*}[!t]
\centering
\subfloat[MSD]{\includegraphics[width=3.3in]{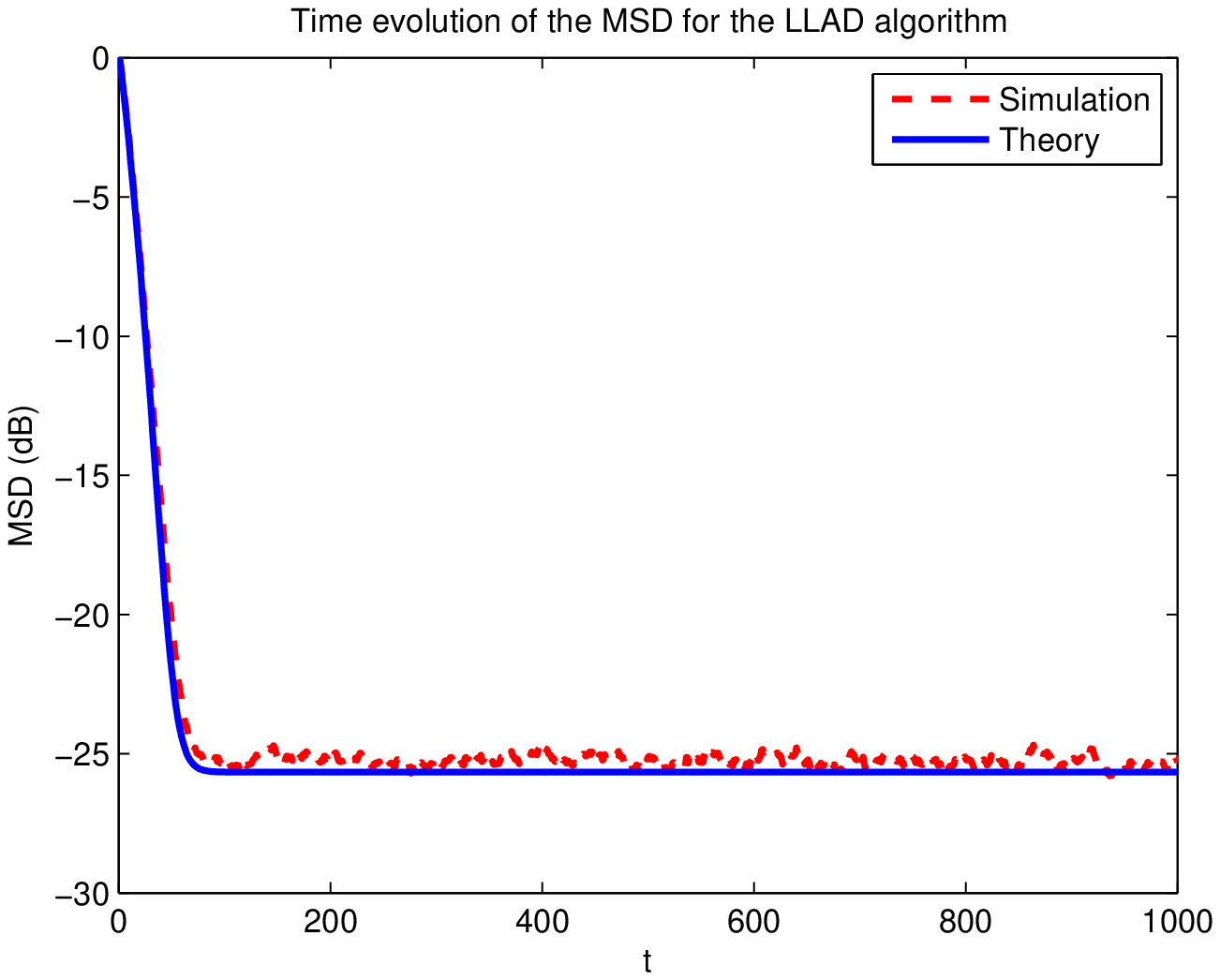}}
\hfil
\subfloat[EMSE]{\includegraphics[width=3.3in]{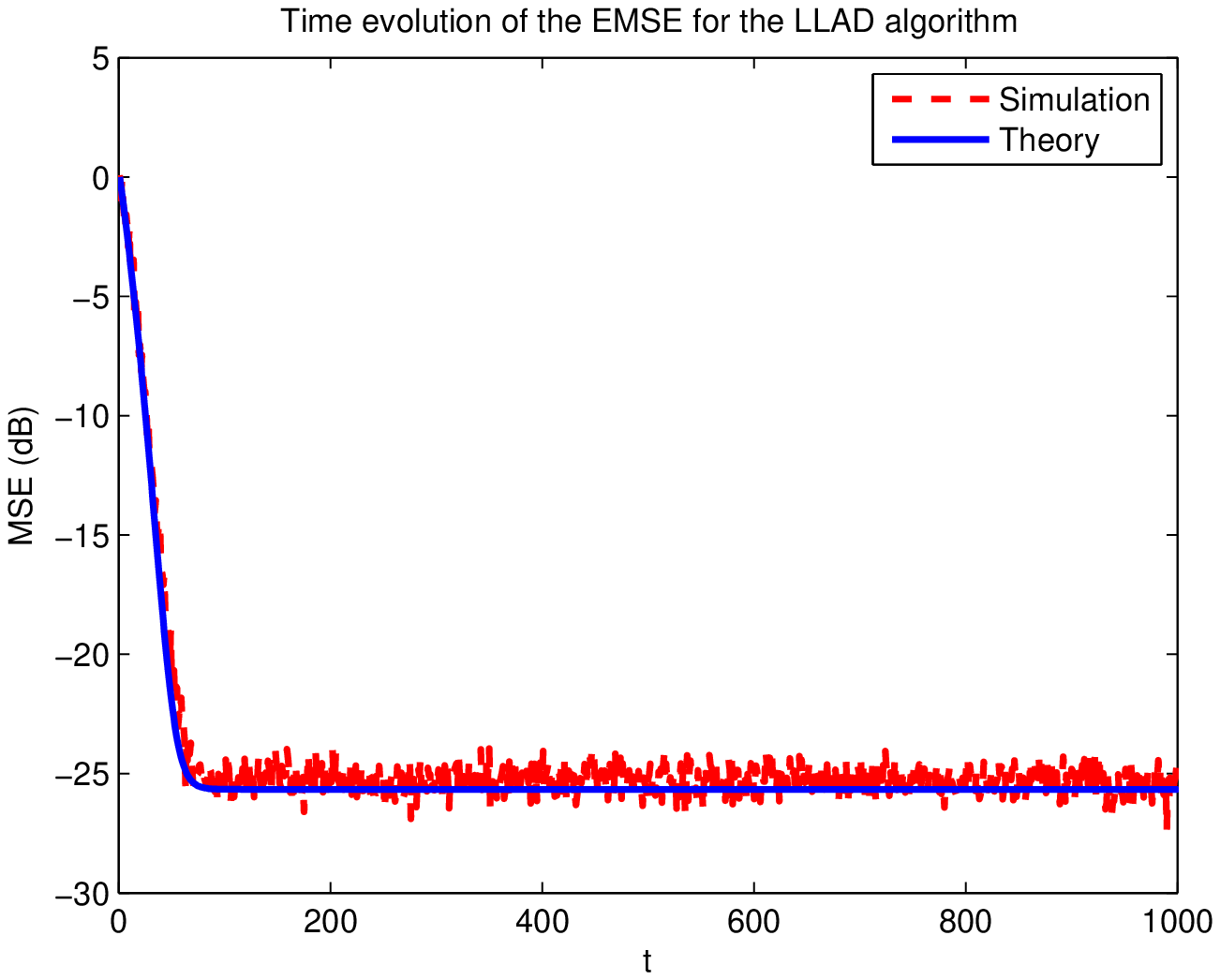}}
\caption{Theoretical and simulated MSD and EMSE for the LLAD algorithm.}
\label{fig_transient_LLAD}
\end{figure*}

\subsection{Steady State Analysis}
At the steady state,~\eqref{energy1} and~\eqref{eq:2} yields
\begin{align}
\mu E\left[\|\vx_t\|_{\mSig}^2\right]h_U(e_t) = 2\,h_G(e_t)E\left[e_{a,t}^{\mSig}e_{a,t}\right].\label{eq:steady}
\end{align}
Without loss of generality, we set the weight matrix $\mSig = \mI$, then~\eqref{eq:steady} leads the steady state EMSE
\begin{align}
  \zeta & \defi E[e_{a,t}^2] \nn\\
        & = \frac{\mu}{2}E\left[\|\vx_t\|^2\right]\frac{h_U(e_t)}{h_G(e_t)} \nn\\
        & = \frac{\mu}{2}\Tr(\mR)\frac{h_U(e_t)}{h_G(e_t)}.\label{eq:EMSE}
\end{align}
By Assumption 1, the steady state MSD is given by~\cite{tareq2003}
\begin{align*}
  \eta & \defi E\left[\|\vtw_t\|^2\right] \nn\\
       & = \frac{p}{\Tr(\mR)} \zeta, \nn
\end{align*}
where $p$ denotes the filter length.

At the steady state, we additionally use the following assumptions, which directly follow from the property of a learning algorithm that as $t$ goes to infinity, $e_t$ goes to zero.

\begin{description}
  \item[{\it Assumption 5:}] \hfill \\
For sufficiently small $\mu$, $h_G(e_t)$ and $h_U(e_t)$ functions of the LMLS algorithm as $t \rightarrow \infty$ is given by
\begin{align*}
&h_G(e_t) = \frac{1}{\sigma_e^2}E\left[\frac{\alpha e_t^4}{1+\alpha e_t^2}\right] \rightarrow \frac{\alpha}{\sigma_e^2}E\left[e_t^4\right],\\
&h_U(e_t) = E\left[\frac{\alpha^2e_t^6}{\left(1+\alpha e_t^2\right)^2}\right] \rightarrow \alpha^2 E\left[e_t^6\right].
\end{align*}
   \item[{\it Assumption 6:}]\hfill\\
For sufficiently small $\mu$, $h_G(e_t)$ and $h_U(e_t)$ functions of the LLAD algorithm as $t \rightarrow \infty$ is given by
\begin{align*}
&h_G(e_t) = \frac{1}{\sigma_e^2}E\left[\frac{\alpha e_t^2}{1+\alpha |e_t|}\right] \rightarrow \frac{\alpha}{\sigma_e^2}E\left[e_t^2\right],\\
&h_U(e_t) = E\left[\frac{\alpha^2e_t^2}{\left(1+\alpha|e_t|\right)^2}\right]\rightarrow\alpha^2E\left[e_t^2\right].
\end{align*}
\end{description}

Now, we explicitly derive the steady state analysis of the LMLS and LLAD algorithms, respectively.

\noindent
{\it The LMLS Algorithm:} For the LMLS algorithm, by Assumption 5,~\eqref{eq:EMSE} leads
\begin{align}
\zeta_{\mathrm{LMLS}} = \frac{\mu}{2}\alpha \Tr(\mR)\sigma_e^2\frac{E\left[e^6_t\right]}{E\left[e^4_t\right]}.\label{eq:steady-LMLS2}
\end{align}
By Assumption 2, $e_t$ is a Gaussian random variable and $\sigma_e^2 = \zeta + \sigma_n^2$, we have
\begin{align*}
\zeta_{\mathrm{LMLS}} &= \frac{\mu}{2}\alpha\Tr(\mR)\sigma_e^2\frac{15\sigma_e^6}{3\sigma_e^4},\nn\\
&=\frac{5\mu}{2}\alpha \Tr(\mR)\left(\zeta_{LMLS} + \sigma_n^2\right)^2.
\end{align*}
Hence, after some algebra, the EMSE and MSD for the LMLS algorithm are given by
\begin{align}
&\zeta_{\mathrm{LMLS}} = \frac{1-5\alpha\mu\Tr(\mR)\sigma_n^2 \pm \sqrt{1-10\alpha\mu\Tr(\mR)\sigma_n^2}}{5\alpha\mu\Tr(\mR)},\label{eq:EMSE-LMLS}\\
&\eta_{\mathrm{LMLS}} = p\frac{1-5\alpha\mu\Tr(\mR)\sigma_n^2 \pm \sqrt{1-10\alpha\mu\Tr(\mR)\sigma_n^2}}{5\alpha\mu\Tr(\mR)^2},\nn
\end{align}
where the smaller roots match with the simulations. Note that~\eqref{eq:EMSE-LMLS} for $\alpha = 1$ is the same with the EMSE of the LMF algorithm~\cite{tareq2003}. \\

\noindent {\bf Remark 4.2:}
In~\eqref{eq:EMSE-LMLS}, let $\tilde{\mu} \defi \mu\alpha$, then
\begin{align}
\zeta_{\mathrm{LMLS}} = \frac{1-5\tilde{\mu}\Tr(\mR)\sigma_n^2 \pm \sqrt{1-10\tilde{\mu}\Tr(\mR)\sigma_n^2}}{5\tilde{\mu}\Tr(\mR)}.\label{eq:EMSE-LMLS2}
\end{align}
By~\eqref{eq:EMSE-LMLS2}, we could achieve similar steady state convergence performance for different $\alpha$ by changing the step size $\mu$, e.g., $\tilde{\mu} = \mu\alpha = \frac{\mu}{10}10\alpha$, however, smaller $\alpha$ results in a slower convergence rate. Hence, without loss of generality, we propose the algorithms with $\alpha = 1$ under the Gaussianity assumption.\\

\noindent
{\it The LLAD Algorithm:} Similarly, for the LLAD algorithm, by Assumption 6,~\eqref{eq:EMSE} yields
\begin{align*}
\zeta_{\mathrm{LLAD}} &= \frac{\mu}{2}\Tr(\mR)\sigma_e^2\alpha\frac{E[e^2_t]}{E[e^2_t]},\nn\\
&= \frac{\mu\alpha}{2}\Tr(\mR)\sigma_e^2.\nn
\end{align*}
By Assumption 2, the EMSE and MSD for the LLAD algorithm is given by
\begin{align}
&\zeta_{\mathrm{LLAD}} = \frac{\mu\alpha\Tr(\mR)\sigma_n^2}{2-\mu\alpha\Tr(\mR)}.\label{eq:EMSE-LLAD}\\
&\eta_{\mathrm{LLAD}} = \frac{\mu\alpha p\sigma_n^2}{2-\mu\alpha\Tr(\mR)}\nn
\end{align}
Note that~\eqref{eq:EMSE-LLAD} is the same with the EMSE of the LMS algorithm~\cite{tareq2003}. Hence, for sufficiently small $\alpha$, the LLAD algorithm achieves similar steady-state convergence performance with the LMS algorithm under the zero-mean Gaussian error signal assumption.

In Fig.~\ref{fig_steady}, we plot the theoretical and simulated MSD vs. step size for the LMLS and LLAD algorithms. In the system identification framework, we choose the regressor and noise signals as i.i.d. zero mean Gaussian with the variances $\sigma_x^2 = 1$ and $\sigma_n^2 = 0.01$, respectively. The parameter of interest $\vw_o \in \mathbbm{R}^5$ is randomly chosen. We observe that the theoretical steady-state MSD matches with the simulation results generated through the ensemble average of the last $10^3$ iterations of $10^5$ (for the LMLS algorithm) and $10^4$ (for the LLAD algorithm) iterations of 200 independent trials. In. Fig.~\ref{fig_transient_LMLS} and Fig.~\ref{fig_transient_LLAD}, under the same configurations, we compare the simulated MSD and EMSE curves generated through the ensemble average of 200 independent trials with the theoretical results for the step-size $\mu = 0.1$. We note that theoretical performance analyzes match with our simulation results.

\subsection{Tracking Performance}
In this subsection, we investigate the tracking performance of the introduced algorithms in a non-stationary environment. We assume a random walk model~\cite{sayed_book} for $\vw_{o,t}$ such that
\begin{align}
\vw_{o,t+1} =\vw_{o,t}+\vq_t\label{eq:random}
\end{align}
where $\vq_t \in \Rp$ is a zero-mean vector process with covariance matrix $E[\vq_t\vq_t^T] = \mQ$. We note that the model~\eqref{eq:random} has not changed the definitions of \emph{a priori} error. Hence, by the Assumption 5, the tracking EMSE of the LMLS is the same with the tracking EMSE of the LMF and is approximately given by~\cite{sayed_book}
\begin{align*}
\zeta_{\mathrm{LMLS}}' \approx \frac{3\alpha\mu \sigma_n^4\Tr(\mR)+\mu^{-1}\Tr(\mQ)}{6\sigma_n^2}.
\end{align*}
Similarly, through the Assumption 6, we obtain the tracking EMSE of the LLAD as
\begin{align*}
\zeta_{\mathrm{LLAD}}' = \frac{\alpha\mu\sigma_n^2\Tr(\mR)+\mu^{-1}\Tr(\mQ)}{2-\alpha\mu\Tr(\mR)}.
\end{align*}

In the next section, we compare the new algorithms with the conventional LMS and SA in terms of the stability bound and robustness.

\section{Comparison with the Conventional Algorithms}
We re-emphasize that the cost function $J(e_t)$ intrinsically combines the costs, mainly, $F(e_t)$ and $F^2(e_t)$ based on the relative error amount since for small perturbations on the error, the updates are mainly using the cost $F^2(e_t)$. Based on our stochastic gradient approach, i.e., removing the expectation in the gradient descent, $F^2(e_t)$ and $F(e_t^2)$ results in the same algorithm. Hence, in this section we compare the stability of the LMLS algorithm with the LMF and LMS algorithms and analyze the robustness of the LLAD algorithm in the impulsive noise environments.

\subsection{Stability Bound for the LMLS Algorithm}
We again refer to the stochastic gradient update~\eqref{eq:update}, which we rewrite as
\begin{align*}
\vw_{t+1}=\vw_t + \mu' \vx_t\nabla_{e_t}f(e_t),
\end{align*}
where $\mu' \defi \mu\frac{\alpha f(e_t)}{1+\alpha f(e_t)}$. Note that $\mu' \leq \mu$ irrespective of the design parameter $\alpha$. Hence, intuitively we can state that for the introduced algorithms the step-size bound is at least as large as the step-size bound for the corresponding conventional algorithm.

Analytically, for stable updates the step size $\mu$ should satisfy
\begin{align*}
E\left[\|\vtw_{t+1}\|^2\right] \leq E\left[\|\vtw_t\|^2\right].
\end{align*}
By~\eqref{energy1}, the Assumption 3, and $\mSig = \mI$, the stability bound on the step size is given by
\begin{align*}
\mu \leq \frac{2}{E\left[\|\vx_t\|^2\right]}\underset{E[e_{a,t}^2]\in\Omega}{\mathrm{inf}}\left\{E[e_{a,t}e_t]\frac{h_G(e_t)}{h_U(e_t)}\right\},
\end{align*}
where
\begin{align*}
\Omega \defi \left\{E[e_{a,t}^2]: \lambda \leq E[e_{a,t}^2] \leq \frac{1}{4}\Tr(\mR)E[\|\vtw_{0}\|^2]\right\},
\end{align*}
with the Cramer-Rao lower bound $\lambda$~\cite{van2004detection}. For example the step size bound for the LMLS yields
\begin{align*}
\mu \leq \frac{1}{E\left[\|\vx_t\|^2\right]}\underset{E[e_{a,t}^2]\in\Omega}{\mathrm{inf}}\left\{\frac{E[e_{a,t}e_t]}{E\left[e^2_t\right]}\beta\right\},
\end{align*}
where
\begin{align*}
\beta \defi& \frac{E\left[\frac{\alpha e^4_t}{1+\alpha e^2_t}\right]}{E\left[\frac{\alpha^2 e^6_t}{\left(1+\alpha e^2_t\right)^2}\right]} \\
=& \frac{E\left[\frac{\alpha e^4_t}{\left(1+\alpha e^2_t\right)^2}\right]+E\left[\frac{\alpha^2e^6_t}{\left(1+\alpha e^2_t\right)^2}\right]}{E\left[\frac{\alpha^2 e^6_t}{\left(1+\alpha e^2_t\right)^2}\right]} \geq 1.
\end{align*}
We re-emphasize that the LMLS extends the stability bound of the LMS algorithm (the same bound with $\beta=1$) while performing comparable performance with the LMF algorithm, which has several stability issues \cite{nascimento2005conf,nascimento2006,hubscher2007}.

\subsection{Robustness Analysis for the LLAD Algorithm}
Although the performance analysis of the adaptive filters assumes the white Gaussian noise signals, in practical applications the impulsive noise is a common problem~\cite{shao1993}. In order to analyze the performance in the impulsive noise environments, we use the following model.\\

\noindent
{\bf Impulsive noise model:}
We model the noise as a summation of two independent random terms~\cite{wang1997,chan2004} as
\begin{align*}
n_t = n_{o,t} + b_tn_{i,t},
\end{align*}
where $n_{o,t}$ is the ordinary noise signal that is zero-mean Gaussian with variance $\sigma_{n_o}^2$ and $n_{i,t}$ is the impulse-noise that is also zero-mean Gaussian with significantly large variance $\sigma_{n_i}^2$. Here, $b_t$ is generated through a Bernoulli random process and determines the occurrence of the impulses in the noise signal with $p_B(b_t=1) = \nu_i$ and $p_B(b_t=0)=1-\nu_i $ where $\nu_i$ is the frequency of the impulses in the noise signal. The corresponding probability density function is given by
\begin{align*}
p_n(n_t) = \frac{1-\nu_i}{\sqrt{2\pi}\sigma_{n_o}}\mathrm{exp}\left(-\frac{n^2_t}{2\sigma_{n_o}^2}\right) + \frac{\nu_i}{\sqrt{2\pi}\sigma_{n}}\mathrm{exp}\left(-\frac{n^2_t}{2\sigma_{n}^2}\right),
\end{align*}
where $\sigma_n^2 = \sigma_{n_o}^2 + \sigma_{n_i}^2$.\\

We particularly analyze the steady-state performance of the LLAD algorithm (for which $f(e_t) = |e_t|$) in the impulsive noise environments since we motivate the LLAD algorithm as improving the steady state convergence performance of the SA.
Since the noise is not a Gaussian random variable in impulsive noise environment, the Gaussianity assumption of the estimation error $e_t$ and the Price's Theorem are not applicable. At the steady-state, for $\mSig = \mI$, \eqref{energy1} yields
\begin{align}
E\left[\|\vx_t\|^2\right] = \frac{2 E\left[\frac{\alpha e_{a,t}e_t}{1+\alpha|e_t|}\right]}{\mu E\left[\frac{\alpha^2e^2_t}{\left(1+\alpha|e_t|\right)^2}\right]}.\label{eq:imp-steady}
\end{align}
We now evaluate the each term in \eqref{eq:imp-steady} separately. We first consider the nominator of the RHS of \eqref{eq:imp-steady}, and write
{\small \begin{align}
    & E\left[\frac{\alpha e_{a,t}e_t}{1+\alpha|e_t|}\right] \nn\\
    & = \int_{-\infty}^{\infty}\int_{-\infty}^{\infty}\frac{\alpha e_{a,t}(e_{a,t}+n_t)}{1+\alpha|e_{a,t}+n_t|}\frac{\mathrm{exp}\left(-\frac{e_{a,t}^2}{2\sigma_{e_a}^2}\right)}{\sqrt{2\pi}\sigma_{e_a}}p_n (n_t)de_{a,t}dn_t. \nn\\
    & = \alpha\int_{-\infty}^{\infty}\int_{-\infty}^{\infty} e_{a,t}e_t\frac{\mathrm{exp}\left(-\frac{e_{a,t}^2}{2\sigma_{e_a}^2} - \frac{n^2_t}{2\sigma_{n_o}^2}\right)}{2\pi\sigma_{e_a}\sigma_{n_o}}(1-\nu_i)de_{a,t}dn_t \nn\\
    & \hspace{0.1cm} + \int_{-\infty}^{\infty}\int_{-\infty}^{\infty} e_{a,t}\sign(e_{a,t}+n_t)\frac{\mathrm{exp}\left(-\frac{e_{a,t}^2}{2\sigma_{e_a}^2} - \frac{n^2_t}{2\sigma_n^2}\right)}{2\pi\sigma_{e_a}\sigma_n}\nu_ide_{a,t}dn_t, \nn
\end{align}}

\noindent where in the last step of the equation we assume that in the impulse-free environment, $\frac{\alpha e_{a,t}e_t}{1+\alpha |e_t|} \approx \alpha e_{a,t} e_t$ since at steady state, the error is assumed to take relatively small values whereas if the impulse-noise occurs, $\frac{\alpha e_{a,t}e_t}{1+\alpha |e_t|} \approx e_{a,t}\sign(e_t)$ due to the large perturbation on the error. Hence, since $\sigma_n^2 \gg \sigma_{e_a}^2$, the expectation leads
\begin{align}
E\left[\frac{\alpha e_{a,t}e_t}{1+\alpha|e_t|}\right] = \alpha(1-\nu_i)\sigma_{e_a}^2+\sqrt{\frac{2}{\pi}}\nu_i\frac{\sigma_{e_a}^2}{\sigma_n}.\label{first-exp}
\end{align}
Following similar steps for the denominator of the RHS of \eqref{eq:imp-steady}, we obtain
\begin{align}
E\left[\frac{\alpha^2e^2_t}{\left(1+\alpha|e_t|\right)^2}\right] = \alpha^2(1-\nu_i)\left(\sigma_{e_a}^2+\sigma_{n_o}^2\right)+\nu_i.\label{second-exp}
\end{align}
By~\eqref{eq:imp-steady},~\eqref{first-exp} and~\eqref{second-exp}, the EMSE of the LLAD algorithm in the impulsive noise environment is given by
\begin{align}
\zeta_{\mathrm{LLAD}}^{*} = \frac{\mu\Tr(\mR)\left(\nu_i + \alpha^2(1-\nu_i)\sigma_{n_o}^2\right)}{\alpha(1-\nu_i)(2-\alpha\mu\Tr(\mR))+\sqrt{\frac{8}{\pi}}\frac{\nu_i}{\sigma_n}}.\label{eq:EMSE-IMP}
\end{align}
Note that for $\nu_i = 0$ (impulse-free)~\eqref{eq:EMSE-IMP} yields~\eqref{eq:EMSE-LLAD}.\\

\begin{figure}[t!]
\centering
\includegraphics[width=3.3in]{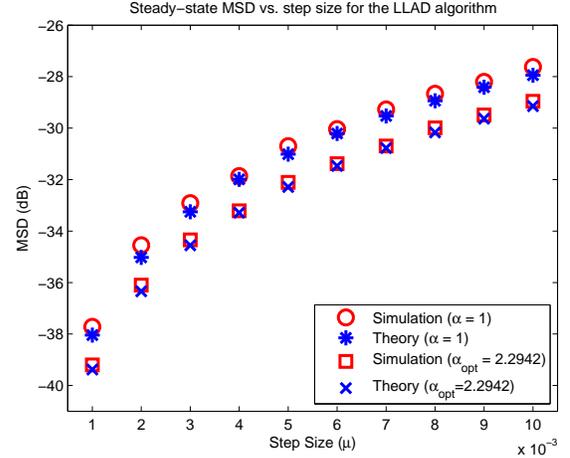}
\caption{Dependence of the steady-state MSD on the step size $\mu$ for the LLAD algorithm in the 5\% impulsive noise environment.}
\label{steady_state_imp}
\end{figure}
\noindent
{\bf Remark 5.1:}
Increasing $\nu_i$ or in other words more frequent impulses cause larger steady state EMSE. However, through the optimization of $\alpha$, we can minimize the steady state EMSE. After some algebra, the optimum design parameter in impulsive noise environment is roughly given by
\begin{align*}
\alpha_{\mathrm{opt}} \approx \sqrt{\frac{\nu_i}{1-\nu_i}}\frac{1}{\sigma_{n_o}}.
\end{align*}

In Fig.~\ref{steady_state_imp}, we plot the dependence of the steady-state MSD with the step size in 5\%, i.e., $\nu_i = 0.05$, impulsive noise environment where $\sigma_x^2 = 1$, $\sigma_{n_o}^2 = 0.01$ and  $\sigma_{n_i}^2 = 10^4$ after 200 independent trials. We observe that $\alpha_{\mathrm{opt}}$ improves the convergence performance and the theoretical analyzes through the impulsive noise model matches with the simulation results. We next demonstrate the performance of the introduced algorithms in different applications.

\section{Numerical Examples}
In this section, we particularly compare the convergence rate of the algorithms for the same steady state MSD through the specific choice of the step sizes for a fair comparison. Here, we have a stationary data $d_t = \vw_o^T\vx_t+n_t$ where $\vx_t$ is zero-mean Gaussian i.i.d. regression signal with variance $\sigma_x^2 = 1$, $n_t$ represent zero-mean i.i.d. noise signal and the parameter of interest $\vw_o \in \R^5$ is randomly chosen. In following scenarios, we compare the algorithms under Gaussian noise and impulsive noise models subsequently.

\begin{figure}[t!]
\centering
\includegraphics[width=3.3in]{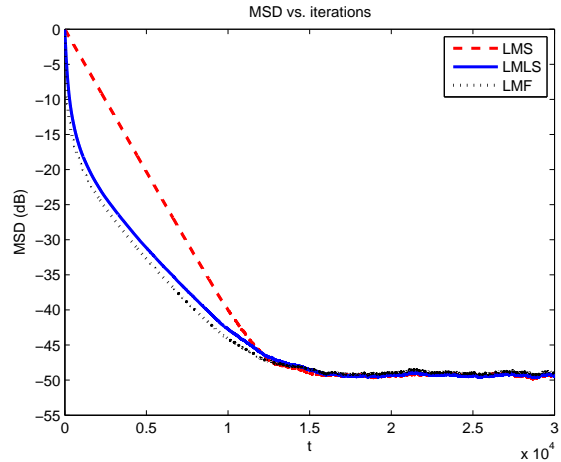}
\caption{Comparison of the MSD of the LMLS, LMS and LMF algorithms for the same steady state MSD where $\mu_{\mathrm{LMLS}} = \mu_{\mathrm{LMF}}= 0.01$ and $\mu_{\mathrm{LMS}} = 0.00047$.}
\label{comp_MSD_LMLS}
\end{figure}

\begin{figure}[t!]
\centering
\includegraphics[width=3.3in]{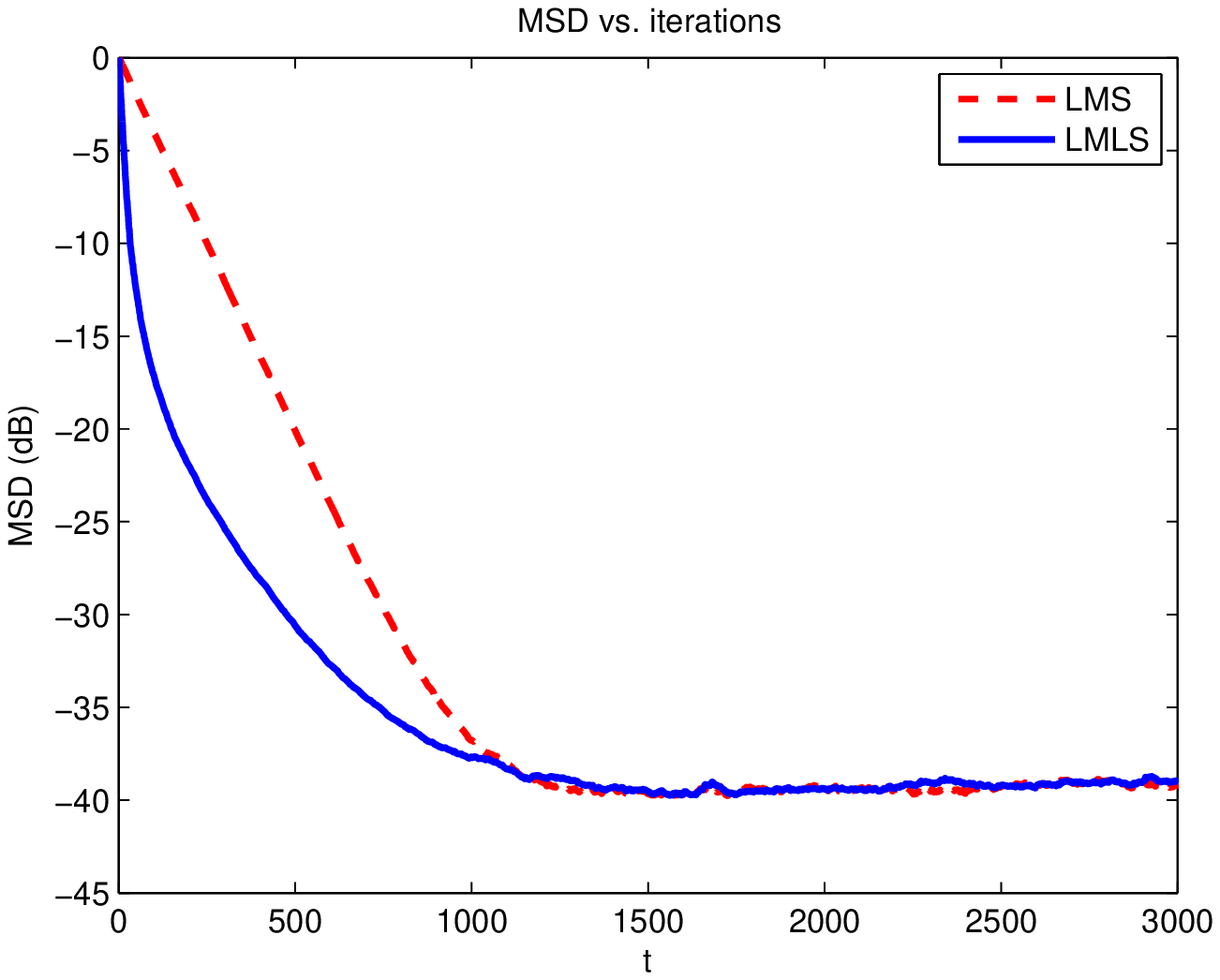}
\caption{Comparison of the MSD of the LMLS and LMS algorithms for the same steady state MSD where $\mu_{\mathrm{LMLS}} = 0.1$ and $\mu_{\mathrm{LMS}} = 0.0047$.}
\label{comp_MSD_LMLS2}
\end{figure}

{\it Scenario 1 (impulse-free environment):}\\
In that scenario, we use a zero-mean Gaussian i.i.d. noise signal with the variance $\sigma_n^2 = 0.01$ and the design parameter $\alpha = 1$. In Fig.~\ref{comp_MSD_LMLS}, we compare the convergence rate of the LMLS, LMF and LMS algorithms for relatively small step sizes. We observe that LMLS and LMF algorithms achieve comparable performance and LMLS achieves better convergence performance than the LMS algorithm. In Fig.~\ref{comp_MSD_LMLS2}, we compare the LMLS and LMS algorithms for relatively large step sizes, i.e., $\mu_{\mathrm{LMLS}} = 0.1$ and $\mu_{\mathrm{LMS}} = 0.0047$. We only compare the LMLS and LMS algorithms since the LMF algorithm is not stable for such a step-size. Hence, the LMLS algorithm demonstrate comparable convergence performance with the LMF algorithm with extended stability bound.

\begin{figure}[t!]
\centering
\includegraphics[width=3.3in]{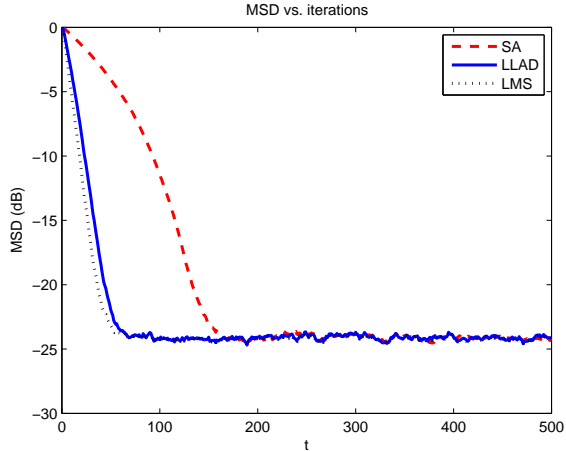}
\caption{Comparison of the MSD of the LLAD, SA and LMS algorithms in impulse-free noise environment with $\mu_{\mathrm{LLAD}}=0.12$, $\mu_{\mathrm{SA}} = 0.01$ and $\mu_{\mathrm{LMS}} = 0.1$.}
\label{comp_MSD_LLAD}
\end{figure}

In Fig.~\ref{comp_MSD_LLAD}, we compare the LLAD, SA and LMS algorithms in impulse-free noise environment. We observe that the LLAD algorithm shows comparable convergence performance with the LMS algorithm, in other words, the logarithmic error cost framework improves the convergence performance of the SA.

\begin{figure*}[!t]
\centering
\subfloat[1\%]{\includegraphics[width=2.4in]{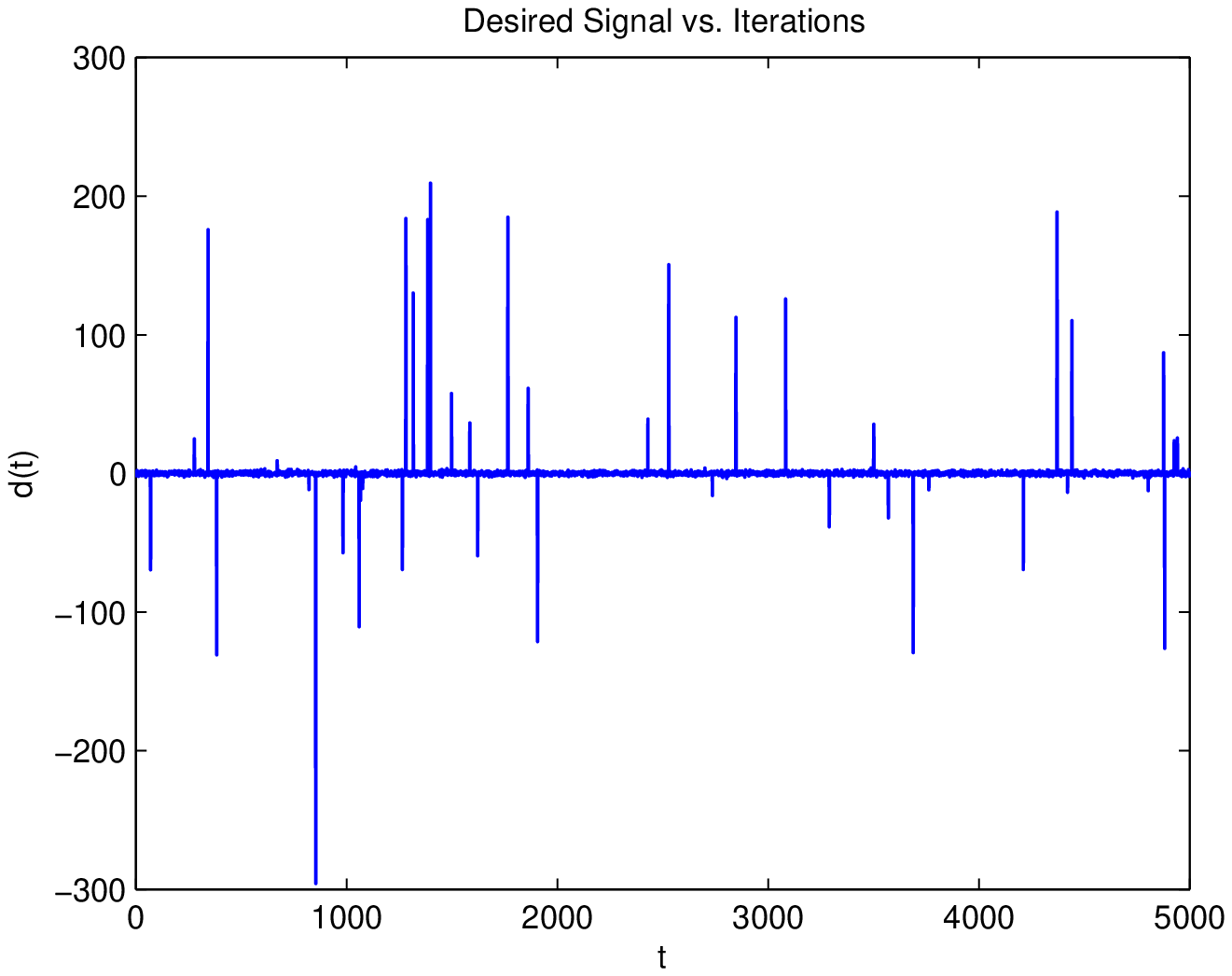}
\label{fig_first}}
\subfloat[2\%]{\includegraphics[width=2.4in]{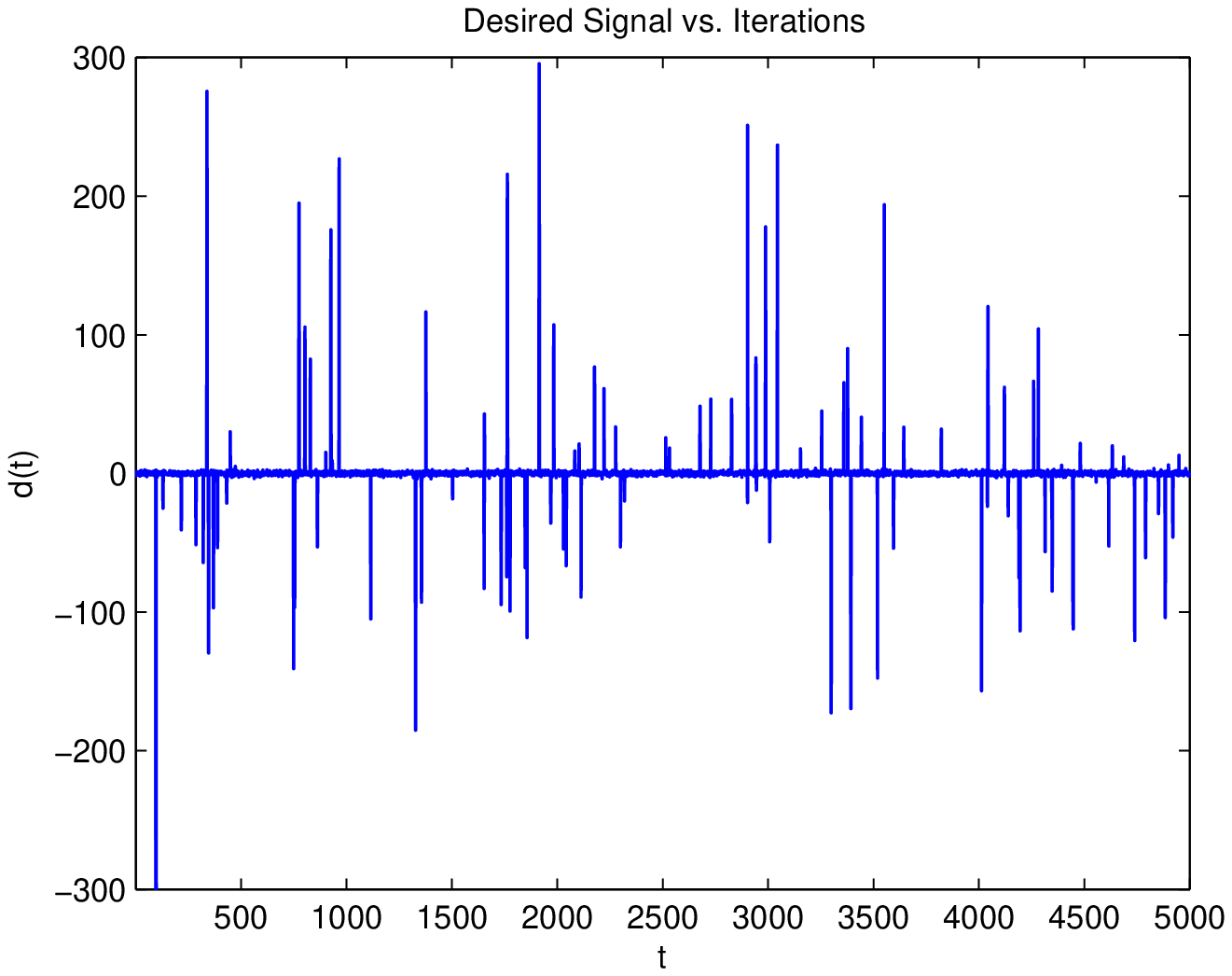}
\label{fig_second}}
\subfloat[5\%]{\includegraphics[width=2.4in]{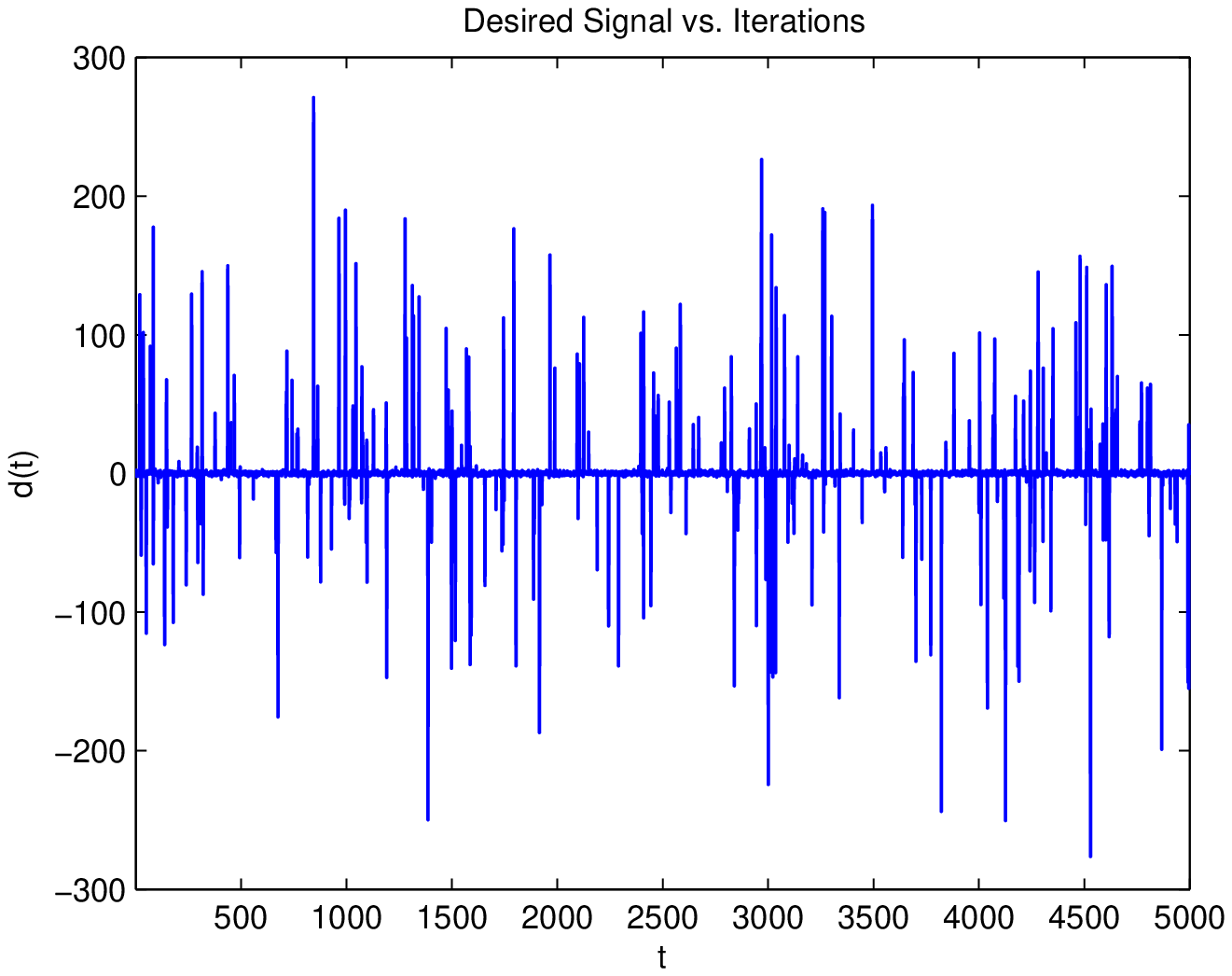}
\label{fig_third}}
\caption{Desired signal in 1\%, 2\% and 5\% impulsive noise environments.}
\label{fig_impulses}
\end{figure*}

\begin{figure*}[!t]
\centering
\subfloat[1\% ($\alpha_{opt} = 1.005$)]{\includegraphics[width=2.4in]{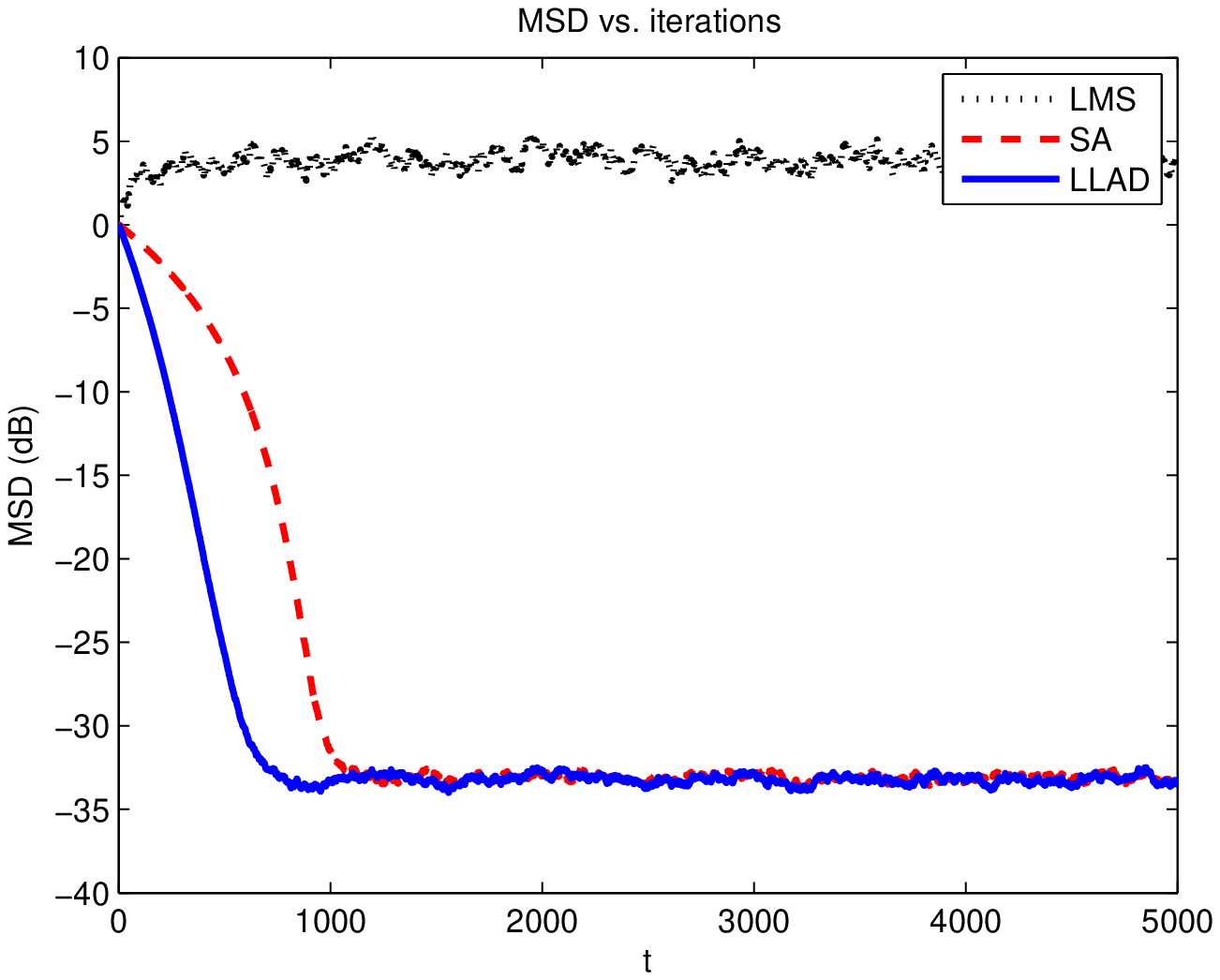}
\label{impulse1}}
\subfloat[2\% ($\alpha_{opt} = 1.4286$)]{\includegraphics[width=2.4in]{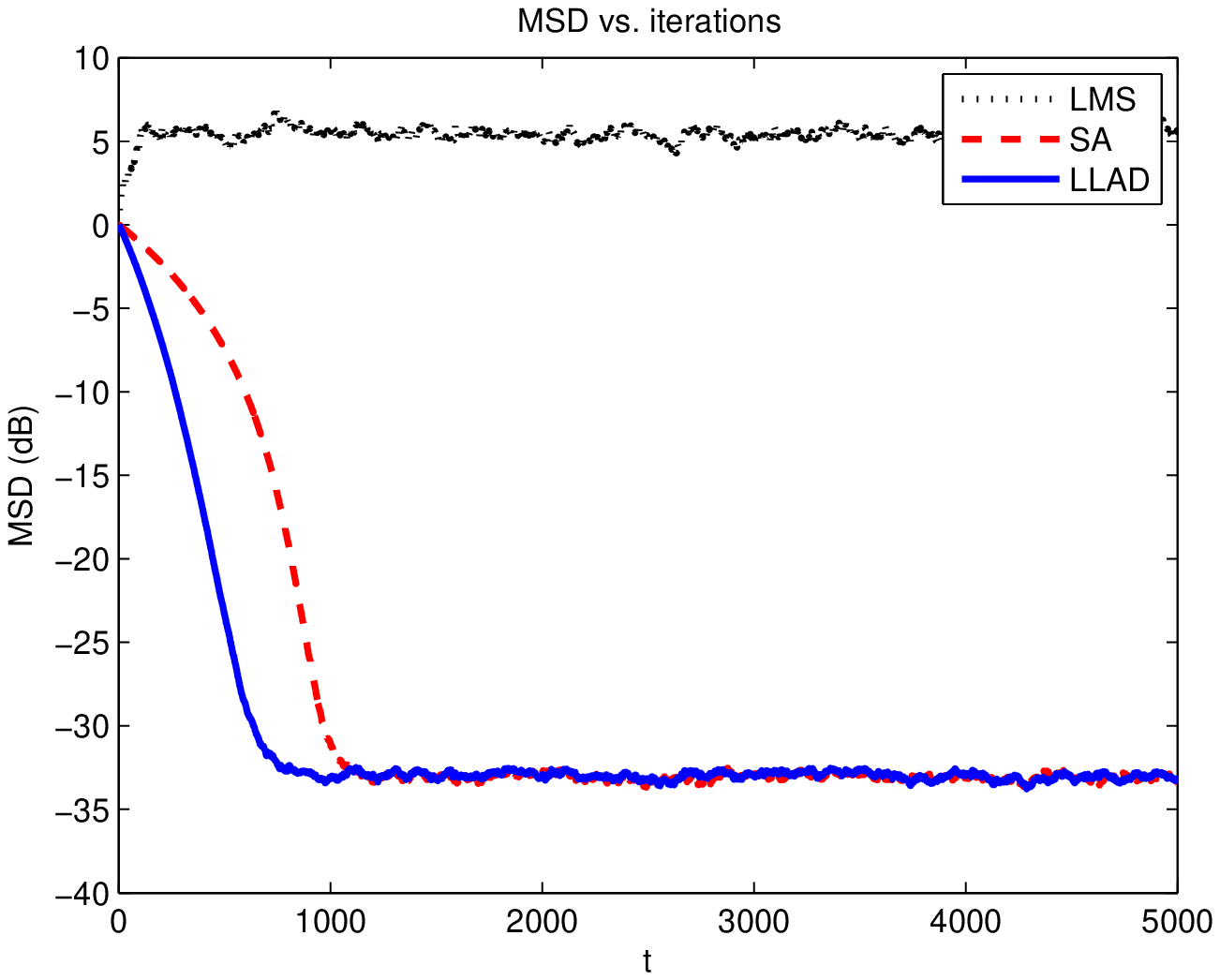}
\label{impulse2}}
\subfloat[5\% ($\alpha_{opt} = 2.2942$)]{\includegraphics[width=2.4in]{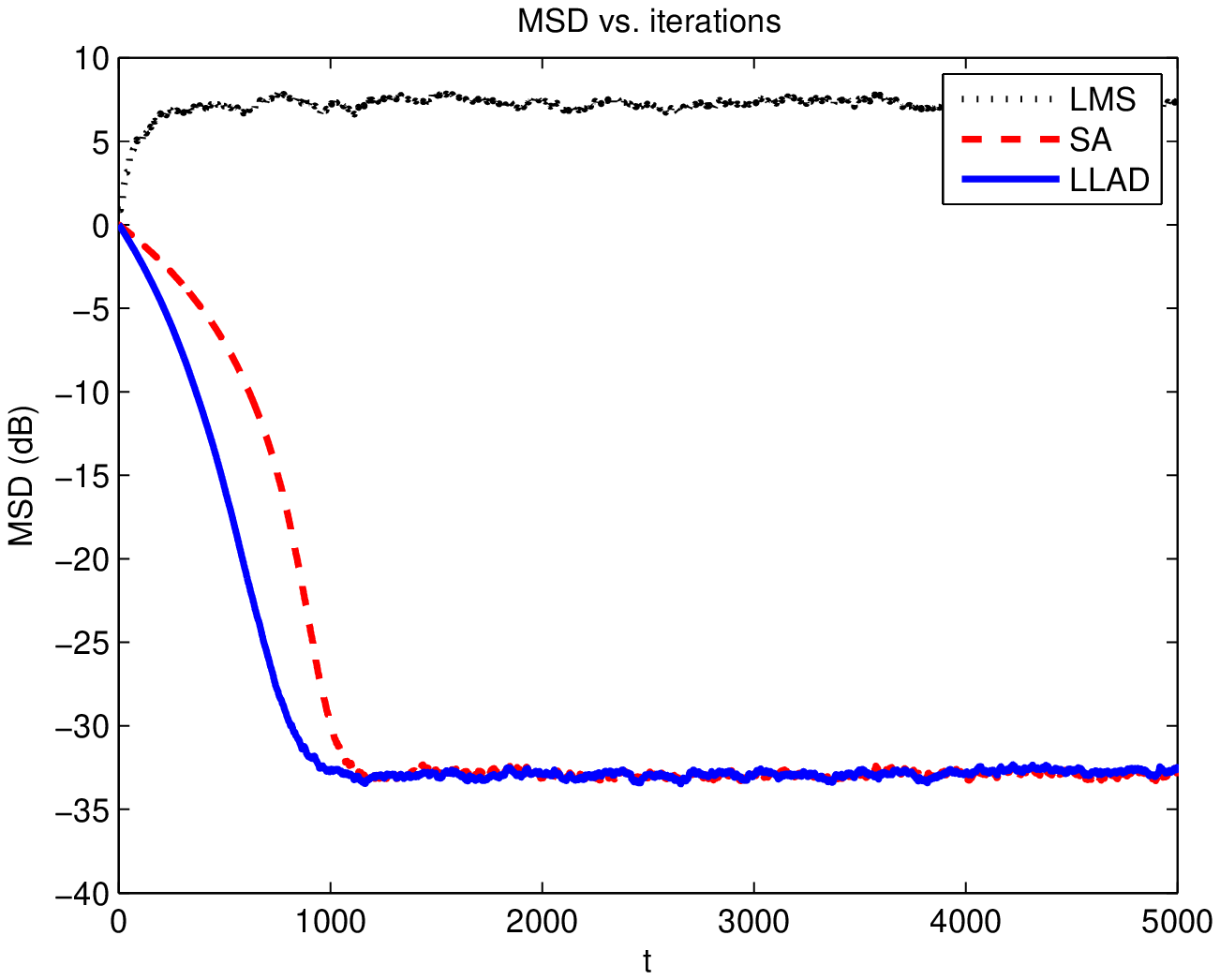}
\label{impulse3}}
\caption{Comparison of the MSD of the LLAD, SA and LMS algorithms in 1\%, 2\% and 5\% impulsive noise environments.}
\label{fig_MSD_imp}
\end{figure*}

{\it Scenario 2 (impulsive noise environment):}\\
Here, we use the impulsive noise model with $\sigma_{n_i}^2 = 10^4$. In that configuration, we resort to the design parameter since through the optimization of $\alpha$, the LLAD algorithm could achieve smaller steady-state MSD. In Fig.~\ref{fig_impulses}, we plot sample desired signals in 1\%, 2\% and 5\% impulsive noise environments and Fig.~\ref{fig_MSD_imp} shows the corresponding time evolution of the MSD of the LLAD, SA and LMS algorithms. The step sizes are chosen as $\mu_{\mathrm{LLAD}} = \mu_{\mathrm{LMS}} = 0.0097, 0.007, 0.0043$ for 1\%, 2\% and 5\% impulsive noise environments, respectively, and $\mu_{\mathrm{SA}} = 0.0015$. The figures show that  in the impulsive noise environments, the LMS algorithm does not converge while the LLAD algorithm, which achieves comparable convergence performance with the LMS algorithm in the impulse free environment, performs still better than the SA.

\section{Concluding Remarks}
In this paper, we present a novel family of adaptive filtering algorithms based on the logarithmic error cost framework. We propose important members of the new family, i.e., the LMLS and LLAD algorithms. The LMLS algorithm achieves comparable convergence performance with the LMF algorithm with far larger stability bound on the step size. In the impulse-free environment, the LLAD algorithm has a similar convergence performance with the LMS algorithm. Furthermore, the LLAD algorithm is robust against impulsive interferences and outperforms the SA. We also provide comprehensive performance analyzes of the introduced algorithms, which match with our simulation results. For example, the steady-state analyzes in the impulse-free and impulsive noise environments. Finally, we show the improved convergence performance of the new algorithms in several different system identification scenarios.

\appendices
\section{Evaluation of $h_G(e_t)$}
{\em The LMLS algorithm:} We have
\begin{align}\label{eq:apndx1}
  h_G(e_t) & = \frac{1}{\sigma_e^2} E\left[ \frac{\alpha e_t^4}{1+\alpha e_t^2} \right], \nn\\
           & = \frac{1}{\sigma_e^2} \left( \sigma_e^2 - \alpha^{-1} + \alpha^{-1} E\left[ \frac{1}{1+\alpha e_t^2} \right] \right),
\end{align}
where $\sigma_e^2 = E[e^2_t]$ and the first line of the equation follows according to the definition of $g(e_t)$ in \eqref{eq:g_func}. According to Assumption 2, we obtain the last term in \eqref{eq:apndx1} as follows
\begin{align}\label{eq:apndx2}
  E\left[\frac{1}{1+\alpha e^2_t}\right] & = \frac{1}{\sqrt{2\pi}\sigma_e}\int_{-\infty}^{\infty} \frac{1}{1+\alpha e^2_t}\mathrm{exp}\left(-\frac{e^2_t}{2\sigma_e^2}\right)de_t \nn\\
    & = \frac{1}{\sqrt{2\alpha\pi}\sigma_e}\int_{-\infty}^{\infty}\frac{\mathrm{exp}\left(-\lambda u^2\right)}{1+u^2}du \nn\\
    & = \frac{1}{\sqrt{2\alpha\pi}\sigma_e} \pi\mathrm{exp}(\lambda)\mathrm{erfc}(\sqrt{\lambda}),
\end{align}
where $u \defi \sqrt{\alpha}e_t$, $\lambda \defi \frac{1}{2\alpha\sigma_e^2}$, and the third line follows from \cite{integrale} with $\mathrm{erfc}(\cdot)$ denoting the complementary error function. Hence, putting \eqref{eq:apndx2} in \eqref{eq:apndx1}, we obtain $h_G(e_t)$ for the LMLS update
\begin{align*}
h_G(e_t) = 1 - 2\lambda\left(1-\sqrt{\pi\lambda}\mathrm{exp}(\lambda)\mathrm{erfc}(\sqrt{\lambda})\right).
\end{align*}

{\em The LLAD algorithm:} We have
\begin{align}\label{eq:apndx3}
  h_G(e_t) & = \frac{1}{\sigma_e^2} E\left[ \frac{\alpha e_t^2}{1+\alpha |e_t|} \right], \nn\\
           & = \frac{1}{\sigma_e^2} \left( E[|e_t|]- \alpha^{-1} + \alpha^{-1} E\left[ \frac{1}{1+\alpha |e_t|} \right] \right),
\end{align}
where the first line follows according to the definition of $g(e_t)$ in \eqref{eq:g_func}. According to Assumption 2, we obtain the last term in \eqref{eq:apndx3} as follows
\begin{align}\label{eq:apndx4}
  E\left[\frac{1}{1+\alpha|e_t|}\right] & = \frac{1}{\sqrt{2\pi}\sigma_e}\int_{-\infty}^{\infty} \frac{1}{1+\alpha|e_t|}\mathrm{exp}\left(-\frac{e^2_t}{2\sigma_e^2}\right)de_t \nn\\
    & = \frac{1}{\sqrt{2\pi}\alpha\sigma_e}\int_{-\infty}^{\infty}\frac{1}{1+|u|}\mathrm{exp}\left(-\kappa u^2\right) du \nn\\
    & = \frac{1}{\sqrt{2\pi}\alpha\sigma_e} \frac{\pi\mathrm{erfi}(\sqrt{\kappa})-\mathrm{Ei}(\kappa)}{\exp(\kappa)},
\end{align}
where $u \defi \alpha e_t$, and $\kappa \defi \frac{1}{2\alpha^2\sigma_e^2}$, and the third line follows from \cite{integrale} with $\mathrm{erfi}(z) = -j\mathrm{erf}(jz)$ denoting the imaginary error function and $\mathrm{Ei}(x)$ denoting the exponential integral, i.e.,
\[
\mathrm{Ei}(x)=-\int_{-x}^{\infty}\frac{\mathrm{exp}(-t)}{t}dt.
\]

As a result, putting \eqref{eq:apndx4} in \eqref{eq:apndx3}, we obtain $h_G(e_t)$ for the LLAD update
\begin{align*}
h_G(e_t) = \frac{1}{\sigma_e}\sqrt{\frac{2}{\pi}}\left(1-\sqrt{\kappa\pi} + \kappa\frac{\pi\mathrm{erfi}(\sqrt{\kappa})-\mathrm{Ei}(\kappa)}{\mathrm{exp}(\kappa)}\right).
\end{align*}

\section{Evaluation of $h_U(e_t)$}
{\em The LMLS Algorithm:} We have
\begin{align}
  h_U(e_t) & = E\left[\frac{\alpha^2e^6_t}{\left(1+\alpha e^2_t\right)^2}\right] \nn\\
           & = E\left[ - \alpha^2 \frac{\partial}{\partial\alpha}\left(\frac{e^4_t}{1+\alpha e^2_t}\right) \right] \nn\\
           & = -\alpha^2 \frac{\partial}{\partial\alpha}\left(E\left[ \frac{e^4_t}{1+\alpha e^2_t} \right] \right), \nn
\end{align}
where in the last line we applied the interchange of integration and differentiation property since $\theta(e_t,\alpha) \defi \frac{e^4_t}{1+\alpha e^2_t}$ and $\frac{\partial\theta(e_t,\alpha)}{\partial\alpha}$ are both continuous in $\R^2$. From Appendix A, we obtain
\begin{align}
  h_U(e_t) & = -\alpha^2 \frac{\partial}{\partial\alpha}\left(\alpha^{-1} E\left[ \frac{\alpha e^4_t}{1+\alpha e^2_t} \right] \right) \nn\\
           & \hspace{-0.8cm} = -\alpha^2 \frac{\partial}{\partial\alpha}\left(\alpha^{-1} \sigma_e^2 h_G(e_t) \right) \nn\\
           & \hspace{-0.8cm} = \sigma_e^2\left(1-2\lambda ( \lambda + 2) + \lambda (2 \lambda + 5) \sqrt{\pi\lambda}\mathrm{exp}(\lambda)\mathrm{erfc}\left( \sqrt{\lambda} \right)\right). \nn
\end{align}

{\em The LLAD Algorithm:} Following similar lines to LMLS algorithm, we have
\begin{align}
  h_U(e_t) & = E\left[\frac{\alpha^2e^2_t}{\left(1+\alpha |e_t|\right)^2}\right] \nn\\
           & = E\left[ - \alpha^2 \frac{\partial}{\partial\alpha}\left(\frac{|e_t|}{1+\alpha |e_t|}\right) \right] \nn\\
           & = -\alpha^2 \frac{\partial}{\partial\alpha}\left(E\left[ \frac{|e_t|}{1+\alpha |e_t|} \right] \right), \nn
\end{align}
where in the last line we applied the interchange of integration and differentiation property since $\theta(e_t,\alpha) \defi \frac{|e_t|}{1+\alpha |e_t|}$ and $\frac{\partial\theta(e_t,\alpha)}{\partial\alpha}$ are both continuous in $\R^2$. From Appendix A, we obtain
\begin{align}
  h_U(e_t) & = -\alpha^2 \frac{\partial}{\partial\alpha}\left(\alpha^{-1} E\left[ \frac{\alpha |e_t|}{1+\alpha |e_t|} \right] \right) \nn\\
           & \hspace{-0.3cm} = -\alpha^2 \frac{\partial}{\partial\alpha}\left(\alpha^{-1} \left( 1 - E\left[ \frac{1}{1+\alpha |e_t|} \right] \right) \right) \nn\\
           & \hspace{-0.3cm} = -\alpha^2 \frac{\partial}{\partial\alpha}\left(\alpha^{-1} \left( 1 - \frac{1}{\sqrt{2\pi}\alpha\sigma_e} \frac{\pi\mathrm{erfi}(\sqrt{\kappa})-\mathrm{Ei}(\kappa)}{\exp(\kappa)} \right) \right) \nn\\
           & \hspace{-0.3cm} = 1 - 2\kappa + 2 \sqrt{\frac{\kappa}{\pi}} \left( 1 + (\kappa-1) \frac{\pi \mathrm{erfi}\left(\sqrt{\kappa}\right)-\mathrm{Ei}(\kappa)}{\mathrm{exp}(\kappa)} \right), \nn
\end{align}
where the third line follows from \eqref{eq:apndx4}.

\bibliographystyle{IEEEtran}
\bibliography{my_references}

\end{document}